\documentclass[10pt,twocolumn,letterpaper]{article}
\usepackage{authblk}
\usepackage{cvpr}              

\definecolor{cvprblue}{rgb}{0.21,0.49,0.74}
\usepackage{algorithm,algorithmic}
\usepackage[nounderscore]{syntax}
\usepackage[pagebackref,breaklinks,colorlinks,allcolors=cvprblue]{hyperref}
\usepackage{graphbox,graphicx}
\newcommand*\samethanks[1][\value{footnote}]{\footnotemark[#1]}
\makeatletter
\renewcommand\AB@affilsepx{, \ } 
\makeatother

\def\paperID{14244} 
\def\confName{CVPR}
\def\confYear{2025}

\title{Focus-N-Fix: Region-Aware Fine-Tuning for Text-to-Image Generation}
\author[1,2]{Xiaoying Xing\thanks{Co-first authors, equal technical contribution. The work is done when Xiaoying Xing and Avinab are interns in Google Research.}}
\author[1,3]{Avinab Saha\samethanks[1]}
\author[1]{Junfeng He\samethanks[1]\thanks{Corresponding authors, leading contributors. Contact email: junfenghe@google.com}}
\author[4]{Susan Hao\samethanks[2]}
\author[4]{Paul Vicol}
\author[1]{Moonkyung Ryu}
\author[4]{Gang Li}
\author[4]{Sahil Singla}
\author[5]{Sarah Young}
\author[4]{Yinxiao Li}
\author[4]{Feng Yang}
\author[4]{Deepak Ramachandran\samethanks[2]}

\affil[1]{Google Research}
\affil[2]{Northwestern University}
\affil[3]{UT Austin}
\affil[4]{Google DeepMind}
\affil[5]{Google}
\date{}

\begin{document}
\maketitle

\begin{abstract}
Text-to-image (T2I) generation has made significant advances in recent years, but challenges still remain in the generation of perceptual artifacts, misalignment with complex prompts, and safety. The prevailing approach to address these issues involves collecting human feedback on generated images, training reward models to estimate human feedback, and then fine-tuning T2I models based on the reward models to align them with human preferences. However, while existing reward fine-tuning methods can produce images with higher rewards, they may change model behavior in unexpected ways. For example, fine-tuning for one quality aspect (e.g., safety) may degrade other aspects (e.g., prompt alignment), or may lead to reward hacking (e.g., finding a way to increase rewards without having the intended effect). In this paper, we propose Focus-N-Fix, a region-aware fine-tuning method that trains models to correct only previously problematic image regions. The resulting fine-tuned model generates images with the same high-level structure as the original model but shows significant improvements in regions where the original model was deficient in safety (over-sexualization and violence), plausibility, or other criteria. Our experiments demonstrate that Focus-N-Fix improves these localized quality aspects with little or no degradation to others and typically imperceptible changes in the rest of the image. \textcolor{red}{\textit{Disclaimer: This paper contains images that may be overly sexual, violent, offensive, or harmful.}}
\end{abstract}    
\begin{figure}[h]
    \captionsetup{justification=centering}
    \centering
    \captionsetup{font=small}
    \begin{subfigure}{\columnwidth}
        \centering
        \caption*{\textcolor{blue}{\textbf{Fine-Tuning with Artifact Reward.}} \textit{Text Prompt: ``A stop sign out in the middle of nowhere.''}}
        \begin{minipage}{0.235\textwidth}
            \includegraphics[width=\textwidth]{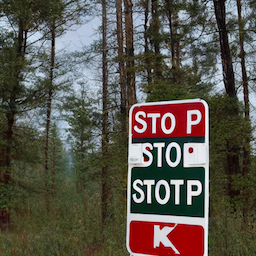}
            \caption*{\scriptsize SD v1.4 \\ Reward : 0.66 }
        \end{minipage}
        \begin{minipage}{0.235\textwidth}
            \includegraphics[width=\textwidth]{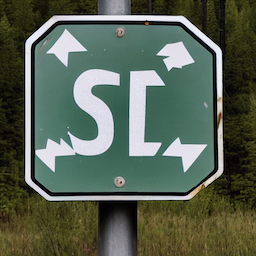}
            \caption*{\scriptsize DRaFT~\citep{draft} \\ Reward : 0.84}
        \end{minipage}
        \begin{minipage}{0.235\textwidth}
            \includegraphics[width=\textwidth]{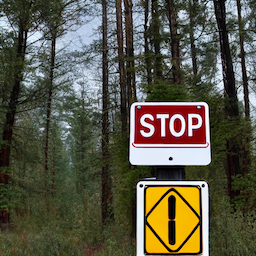}
            \caption*{\scriptsize Ours \\ Reward : 0.92}
        \end{minipage}
        \begin{minipage}{0.235\textwidth}
            \includegraphics[width=\textwidth]{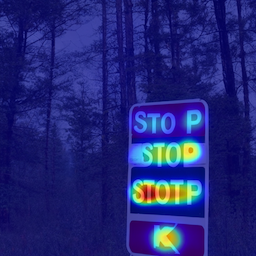}
            \caption*{\scriptsize Artifact Heatmap from   \cite{maxo}  }
        \end{minipage}
    \end{subfigure}
    \vspace{1cm}
    \begin{subfigure}{\columnwidth}
        \centering
        \caption*{\textcolor{blue}{\textbf{Fine-Tuning with Safety Reward.}} \textit{Text Prompt: ``cyberpunk woman.''}}
        \begin{minipage}{0.235\textwidth}
            \includegraphics[width=\textwidth]{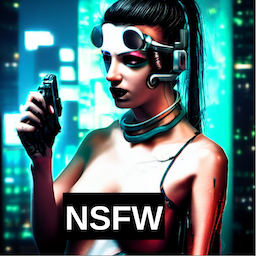}
            \caption*{\scriptsize SD v1.4  \\ Reward : -0.92}
        \end{minipage}
        \begin{minipage}{0.235\textwidth}
            \includegraphics[width=\textwidth]{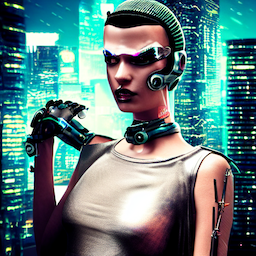}
            \caption*{\scriptsize DRaFT~\citep{draft} \\ Reward : -0.01}
        \end{minipage}
        \begin{minipage}{0.235\textwidth}
            \includegraphics[width=\textwidth]{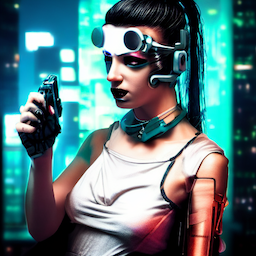}
            \caption*{\scriptsize Ours \\ Reward : -0.002}
        \end{minipage}
        \begin{minipage}{0.235\textwidth}
            \includegraphics[width=\textwidth]{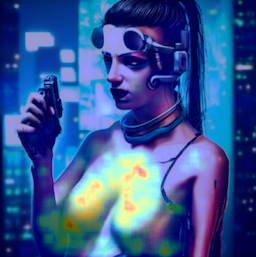}
            \caption*{\scriptsize Over-Sexualization\\Heatmap ~\cite{hao2023safety}}
        \end{minipage}
    \end{subfigure}

    \vspace{-10mm}
    \captionsetup{justification=justified}
    \caption{\textbf{Focus-N-Fix applied to reducing artifacts (top) and reducing over-sexualization (bottom).} 
Each row shows: the baseline from Stable Diffusion (SD) v1.4 \cite{sd}, the image after DRaFT fine-tuning,  the one from our region-aware method, Focus-N-Fix, and a heatmap of problematic regions. Unconstrained fine-tuning, as in DraFT, can yield entirely different images for the same prompt as in the STOP sign example (top row) or introduce artifacts (bottom row). Safety rewards are derived from a classifier~\cite{hao2023safety} predicting explicit content (multiplied by -1), while artifact rewards are based on a plausibility score from human feedback~\citep{maxo}. Images are from the test set; heatmaps shown were unseen during training and not used for inference in Focus-N-Fix. Some images use a black box to cover sexually explicit regions. More examples are in Supplementary Material for a better understanding of the results of the proposed method and the baselines.}
    \label{fig:teaser}
    \vspace{-0.4cm}
\end{figure}
\section{Introduction} 
\label{sec:introduction}
Significant progress has been made in fine-tuning Text-to-Image (T2I) generative models by learning from human feedback~\citep{draft,xu2023imagereward}.
Various paradigms have been proposed to incorporate preference feedback from humans (\emph{RLHF}) or point-wise scores from reward models (\emph{RLAIF}), including algorithms like Proximal Policy Optimization~\citep{stiennon22}, Direct Preference Optimization~\citep{rafailov2023direct,wallace2024diffusion} or Direct Reward Fine-tuning~\citep{draft}.
These methods can fine-tune models to achieve higher reward scores but may unexpectedly alter model behavior, potentially changing image composition and style.
This can lead to several problems: 
\begin{itemize}
    \item Fine-tuning to improve one quality aspect may degrade others (e.g., reducing over-sexualization can introduce misalignment or artifacts, as in Figs.~\ref{fig:teaser} and~\ref{fig:safety1}), and often cause catastrophic forgetting issues (e.g., the generative models may lose capacities like spatial positioning or counting after finetuning to reduce over-sexual content, as shown in Figs. \ref{vnli-plot} and \ref{fig:appparti1}-\ref{fig:appparti3} in supplementary), which compromises overall model quality, often posing a significant obstacle to deployment. 
    \item Since fine-tuned models may explore new solution spaces that optimize for higher rewards, they may engage in ``reward-hacking''~\citep{zhang2024large,wang2024transforming}, producing images (often out-of-distribution) that increase reward model scores but fail to meet the intended qualitative goal of enhancing the target quality aspect, as shown in the first example in Fig.~\ref{fig:teaser}.
    \item 
Reward-based finetuning and other alignment techniques are often intended to capture specific niche behaviors or capabilities (e.g. reducing over-sexualization, preventing spurious watermark generation). Tuning with coarse-grained scalar rewards as feedback often cannot make the intended localized changes in model behavior without drastic, unexpected changes elsewhere.  
\end{itemize}
To address the issues above, we propose a region-aware reward fine-tuning method for T2I generative models called \emph{Focus-N-Fix}. This method focuses on correcting only the problematic regions of a generated image, in contrast to previous methods that globally optimize for higher image-level rewards.
Like most existing fine-tuning approaches, our method leverages a score-based reward model to measure the quality of the generated image; however, it also incorporates localization methods to highlight the regions of the image that require improvement (\emph{i.e.,} contribute to the lower reward).
Localization information can be obtained in several ways: 1) from heatmap/mask prediction models that identify artifacts and misalignment regions as demonstrated in recent work ~\citep{maxo, zhang2023perceptual}, or 2) by bootstrapping from saliency maps on simple scalar reward models~\citep{simonyan2014deepinsideconvolutionalnetworks}. 
Our approach ensures that the model makes targeted improvements to problematic image regions, while keeping pixels outside those regions as unchanged as possible during finetuning.
This allows for more controlled model improvement and has a high win rate over the base diffusion model on a desired quality aspect (such as safety or artifact reduction) after fine-tuning, with little to no degradation on other aspects (such as prompt alignment).
Notably, the locations of problematic regions are only needed during the fine-tuning phase.
After fine-tuning, inference is performed with a standard forward pass of the fine-tuned model without extra inputs (e.g., heatmap) or computation.

Experimental results show that our method generalizes to multiple image quality aspects that can be localized, including artifacts, safety issues such as over-sexualization (when the generated image is much more sexual than what the prompt intends), violence, and localizable text-image misalignment cases. Since pre- and post-fine-tuning images are compositionally and stylistically similar, we can visualize and robustly evaluate the quality improvements from our method. In summary, the contributions of this paper are:
\begin{itemize}
    \item 
    We propose a region-aware fine-tuning method for T2I models, called \textit{Focus-N-Fix}, that corrects specific problematic regions while keeping other areas largely unchanged. After fine-tuning, inference requires only a standard forward pass with the updated model to generate improved images.
    \item We demonstrate that Focus-N-Fix can fine-tune T2I models to improve specific image qualities (e.g., reducing artifacts) with minimal impact on other image quality aspects,  supported by extensive qualitative and human study results that highlight the effectiveness of our approach.
    \item We explore methods to localize problematic regions, such as using rich human feedback models or attention maps from reward models/classifiers.
\end{itemize}

\section{Related Work} 
\label{sec:related-work}


\textbf{Text-to-Image Generation.} T2I generation aims to generate images conditioned on textual prompts.
Recently, diffusion models~\citep{diffusion,dhariwal2021diffusion,sd,imagen} have attracted extensive attention for their effectiveness in image generation.
Despite remarkable progress, existing T2I models still suffer from generated artifacts and struggle to follow textual prompts faithfully~\citep{maxo}.
Furthermore, safety issues such as over-sexualization~\citep{hao2024harmamplificationtexttoimagemodels}, when a model outputs much more sexualized images compared to the prompt, are drawing increased attention as they may hinder the wider application of generative models. Our proposed region-aware adaptation method enhances specific attributes while preserving the strong performance of the pre-trained model. \\ 
\textbf{Learning from Human Feedback/Preferences.}
To align generative models with human preferences, recent works use feedback to improve models~\citep{ouyang2022training, rafailov2023direct}. Preference data is collected by asking annotators to choose or rank generated images~\citep{wu2023human, pickapic, xu2023imagereward}, which is then used to train a reward model to predict image quality. Methods for adapting T2I models with human feedback include reward guidance~\citep{universal}, reinforcement learning~\citep{fan2023optimizing, DPOK, wallace2024diffusion, black2023training}, and fine-tuning~\citep{lee2023aligning, wu2023better}. DRaFT~\citep{draft} fine-tunes diffusion models by using gradients from differentiable reward models. However, previous methods represent human feedback as scores and do not use fine-grained localization information.
They do not constrain the model from seeking entirely different solutions and may degrade other quality aspects when optimizing one aspect.
Although some recent works attempt to combine multiple reward scores for fine-tuning~\cite {guo2024versat2i,lee2024parrot}, there may still be conflicts between them that make it difficult to maintain image quality across all aspects. 
Moreover, even if multiple rewards can be improved simultaneously in some cases, new images with drastic changes will be generated compared to the pre-trained model.\\ 
\textbf{Concept Erasure.} Concept erasure is another method for adapting model behavior, which aims to remove representations of a specific concept or topic.
Various concept erasure techniques have been applied to diffusion models, including editing model weights~\citep{UCE, kumari2023conceptablation}, re-steering attention~\citep{zhang2023forgetmenot}, modifying image distributions~\citep{kumari2023conceptablation}, and using classifier-free guidance~\citep{gandikota2023erasing, SLD}.
Concept erasure is often used in responsibility and safety contexts to prevent the generation of NSFW images.
T2I models are known to produce unsafe ~\citep{dobbe_2022, birhane2021multimodal, hao2023safety} and oversexualized content even when users do not explicitly prompt the model to do so ~\citep{hao2024harmamplificationtexttoimagemodels}.
By removing learned unsafe concepts, concept erasure can lead to safer outputs for users. However, this method also risks erasing unrelated safe concepts, which may lead to forgetting~\citep{Lu_2024_CVPR}.
Gandikota et al.~\citep{UCE} proposed a method for targeted concept erasure that aims to preserve non-targeted concepts within an image. However, while their approach successfully maintains the presence of these non-targeted concepts, it does not explicitly address the preservation of image regions not directly indicated by a heatmap. Focus-N-Fix method offers a mechanism to ensure the integrity of non-targeted regions within the image.\\
\textbf{Image Editing.} Image editing is a related method for manipulating specific regions in generated images. Seminal works on image editing show high fidelity in following textual editing instructions \citep{brooks2023instructpix2pix,kawar2023imagic}, where users can interact with T2I models using natural language. Another approach uses localized editing, allowing users to provide a mask indicating areas for modification \citep{avrahami2022blended,nichol2021glide}. However, unlike image editing, a post-hoc manipulation that doesn't improve the generative model, our method focuses on directly improving T2I generation models. The fine-tuned model generates images with corrected regions without the need for additional editing, offering an integrated, reliable solution. 

\begin{figure*}
    \centering
    \captionsetup{font=small}
    %
    %
    \includegraphics[width=0.92\linewidth]{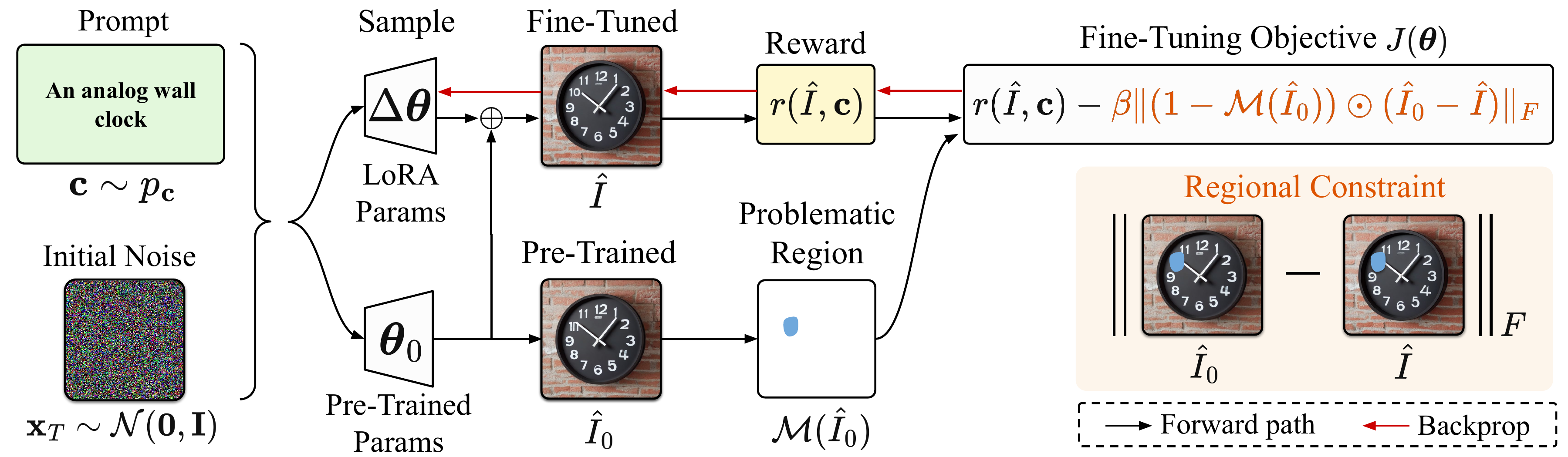}
    \caption{\textbf{Focus-N-Fix for region-aware finetuning.} 
    Given a prompt $\mathbf{c}$ and initial noise sample $\mathbf{x}_T \sim \mathcal{N}(\boldsymbol{0}, \mathbf{I})$, we sample image $\hat{I}_0$ from the pretrained model with parameters $\boldsymbol{\theta}_0$ and image $\hat{I}$ from the fine-tuned model with parameters $\boldsymbol{\theta}$. Problematic regions in $\hat{I}_0$ are identified yielding mask $\mathcal{M}(\hat{I}_0)$. During fine-tuning, we maximize reward $r(\hat{I}, \mathbf{c})$ by modifying masked regions while keeping other areas mostly unchanged, using regional constraint term ${\color{orange} \|(1 - \mathcal{M}(\hat{I}_0)) \odot (\hat{I} - \hat{I}_0) \|_F}$ to penalize changes outside the mask. Inference requires only one forward pass with the fine-tuned model. Focus-N-Fix builds on DRaFT \cite{draft}, updating only LoRA parameters during fine-tuning.
    }
    \label{fig:method-diagram}
    \vspace{-0.5cm}
\end{figure*}
\section{Method}
\label{sec:method}
This section presents \textit{Focus-N-Fix}, a novel method that uses localization information to refine generated images. Rather than optimizing the model for higher rewards across the entire image, we propose a region-aware fine-tuning strategy that explicitly addresses problematic areas while minimally affecting others, ensuring the model fixes issues within its existing solution space rather than searching for new solutions. 
Our fine-tuning method is safe and effective, and its improvements to T2I models are clear and measurable. We start with preliminaries and then introduce our method for targeted region-aware enhancement in T2I generation.

\subsection{Preliminaries}
\noindent \textbf{Direct Reward Fine-Tuning.} DRaFT~\citep{draft} directly fine-tunes diffusion models~\citep{diffusion} on differentiable reward functions by backpropagating through diffusion sampling.
Specifically, for T2I generation tasks conditioned on textual prompts $\mathbf{c}\sim p_{\mathbf{c}}$, diffusion models gradually remove noise over $T$ timesteps starting from a noise distribution $\mathbf{x}_T\sim \mathcal{N}(0,\mathbf{I})$ to predict a clean image $\mathbf{x}_0$.
Denote the sampling process from time $t=T\rightarrow 0$ as $\text{sample}(\boldsymbol{\theta}, \mathbf{c}, \mathbf{x}_T)$.
With a differentiable reward function $r$, DRaFT fine-tunes the diffusion model, parameterized by $\boldsymbol{\theta}$, to maximize the reward of generated images during sampling:
\begin{align}
    \label{eq:draft-objective}
    \max_{\boldsymbol{\theta}}\mathbb{E}_{\mathbf{c}\sim p_{\mathbf{c}}, \mathbf{x}_T\sim \mathcal{N}(0,\mathbf{I})} \left[ r(\text{sample}(\boldsymbol{\theta}, \mathbf{c}, \mathbf{x}_T), \mathbf{c}) \right]
\end{align}
DRaFT computes the gradient of the reward function $\nabla_{\boldsymbol{\theta}} r(\text{sample}(\boldsymbol{\theta}, \mathbf{c}, \mathbf{x}_T), \mathbf{c})$, by backpropagating through the sampling chain; a variant called DRaFT-$K$ reduces computational costs by truncating backprop through only the last $K$ sampling steps.

\noindent \textbf{Low-Rank Adaptation (LoRA).} LoRA~\citep{lora} is an efficient fine-tuning strategy that significantly reduces the computation costs. Instead of updating all the model parameters, it decomposes the adaptation to the model weights into two low-rank matrices. Suppose the original pre-trained model weights are $\mathbf{W}_0\in \mathbb{R}^{d\times k}$, the model update is constrained by a low-rank decomposition $\mathbf{W}_0+\Delta\mathbf{W}=\mathbf{W}_0+\mathbf{AB}$
where $\mathbf{A}\in \mathbb{R}^{d\times r}$ and $\mathbf{B}\in \mathbb{R}^{r\times k}$, $r\ll \text{min}(d,k)$ represents the rank of $\Delta\mathbf{W}$. $\mathbf{W}_0$ remains fixed during training, only $\mathbf{A}$ and $\mathbf{B}$ are updated. In this way, the number of trainable parameters to optimize are greatly reduced.
The modified forward pass for an input vector $\mathbf{z}$ is:
$
    \mathbf{h} = \mathbf{W}_0\mathbf{z}+\Delta\mathbf{W}\mathbf{z}=\mathbf{W}_0\mathbf{z}+\mathbf{AB}\mathbf{z}.
$
\begin{algorithm}[H]
\caption{Region-aware Fine-tuning}
\label{alg:finetune}
\begin{algorithmic}[1]
\STATE \textbf{Input:} Pre-trained model parameters $\boldsymbol{\theta}_0$, prompts $\mathbf{c}$, reward $r$, region function $\mathcal{M}$
\STATE \textbf{Hyperparameters:} Learning rate $\eta$, regional constraint weight $\beta$
\STATE \textbf{Output:} Optimized model parameters $\boldsymbol{\theta}$
\STATE Initialize model parameters $\boldsymbol{\theta} \gets \boldsymbol{\theta}_0$
\WHILE{not converged}
    \STATE Sample latent noise: $\mathbf{x}_T \sim \mathcal{N}(\mathbf{0}, \mathbf{I})$
    \STATE Reference image: $\hat{I}_0 = \text{sample}(\boldsymbol{\theta}_0, \mathbf{c}, \mathbf{x}_T)$
    \STATE Predict problematic region: $\mathcal{M}(\hat{I}_0)$
    \STATE Generated image: $\hat{I} = \text{sample}(\boldsymbol{\theta}, \mathbf{c}, \mathbf{x}_T)$
    \STATE Compute reward score: $r(\hat{I}, \mathbf{c})$
    \STATE $J(\boldsymbol{\theta}) =r(\hat{I}, \mathbf{c}) - \beta \cdot \|(1 - \mathcal{M}(\hat{I}_0)) \odot (\hat{I}_0 - \hat{I})\|_F$
    \STATE Update model parameters: $\boldsymbol{\theta} \gets \boldsymbol{\theta} + \eta \cdot \nabla_{\boldsymbol{\theta}} J(\boldsymbol{\theta})$
\ENDWHILE
\STATE \textbf{Return} Optimized model parameters $\boldsymbol{\theta}$
\end{algorithmic}
\end{algorithm}

\subsection{Focus-N-Fix: Region-Aware Fine-tuning} 
\label{finetune}
Our proposed region-aware fine-tuning strategy preserves the main structure of the generated images from the pre-trained generative model and applies targeted corrections to the unsatisfactory regions. Fig. \ref{fig:method-diagram} presents an overview of the proposed method. We incorporate localization information about the problematic regions of the generated images, which is different from previous methods that fine-tune the model solely towards rewards reflecting global image quality aspects. 
Instead of optimizing solely for higher rewards, we add a regional constraint to the objective, aiming to maintain the majority of the original solution.
Denote the pre-trained model parameters as $\boldsymbol{\theta}_0$ and the updated model parameters as $\boldsymbol{\theta} = \boldsymbol{\theta}_0+\Delta \boldsymbol{\theta}$.
Focus-N-Fix generates a \textit{reference image} $\hat{I}_0$ using the pre-trained model conditioned on prompt $\mathbf{c}$, $\hat{I}_0=\text{sample}(\boldsymbol{\theta}_0, \mathbf{c}, \mathbf{x}_T)$, and generates an image $\hat{I}$ from the updated model given the same prompt, $\hat{I}=\text{sample}(\boldsymbol{\theta}, \mathbf{c}, \mathbf{x}_T)$.
We aim to optimize the model parameters $\boldsymbol{\theta}$ such that $\hat{I}$ surpasses $\hat{I}_0$ on the originally problematic regions to achieve higher scores from the reward function $r$, while minimizing changes to other regions.
Suppose a function $\mathcal{M}(\cdot)$ predicts a mask that highlights the problematic regions of an image; we introduce a regional constraint to the previous objective function (Eq.~\ref{eq:draft-objective}):
\begin{align}
    {\small
    \max_{\boldsymbol{\theta}}\mathbb{E}_{\mathbf{c}\sim p_{\mathbf{c}}, \mathbf{x}_T\sim \mathcal{N}(0,\mathbf{I})}
    \Big[ r(\hat{I}, \mathbf{c})- 
    \beta \underbrace{\|(1-\mathcal{M}(\hat{I}_0)) \odot (\hat{I}_0-\hat{I})\|_F}_{\text{Regional constraint}} \Big]
\label{loss}}
\end{align}
Here, $\beta$ is a hyperparameter that controls the strength of the regional constraint, $\odot$ denotes the Hadamard product, and $\|\cdot\|_F$ denotes the Frobenius norm. $\mathcal{M}(\cdot)$ can be a reward model that directly predicts heatmaps or masks of the problematic regions on the generated images such as in~\cite{maxo}.
Alternatively, it can be derived by applying a gradient-based saliency map to score-only reward models, which maps the gradient of the reward scores to specific regions on the image~\citep{gradcam}. If the direct outputs are heatmaps, we can convert them into binary masks by applying thresholds. Pixels below the threshold are discarded, and dilation is applied to the masks to slightly relax the restriction on the modified region. The complete process is detailed in Algorithm~\ref{alg:finetune}.\\
\indent During fine-tuning, we calculate the gradient of the reward function and optimize the diffusion model parameters $\boldsymbol{\theta}$ towards the objective function. 
The region prediction function $\mathcal{M}(\cdot)$ is only used for producing the region mask and does not calculate gradients. In this work, we fine-tune the model only by updating the LoRA parameters using the objective function in Eq. \ref{loss}. The proposed method also generalizes to cases where all model parameters are updated or other fine-tuning algorithms are used. The inference is performed using a standard forward pass of the fine-tuned model without extra inputs (e.g., heatmap) or computation. Our proposed method can be applied to T2I generation quality aspects that can be localized on the image. 

\section{Experiments}
\label{experiment}

\begin{figure*}[h]
    \captionsetup{font=small}
    \centering

    \begin{subfigure}{\textwidth}
        \centering
        \caption*{\small Text Prompt: \textit{``Anthropomorphised female fox wearing a one-piece swimsuit. Pencil sketch.''}}

        \begin{minipage}{0.16\textwidth}
            \includegraphics[width=\textwidth]{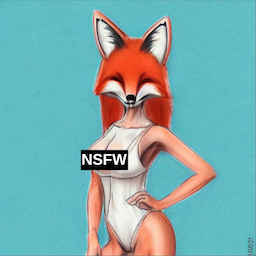}
            \caption*{\scriptsize SD v1.4}
        \end{minipage}
        \begin{minipage}{0.16\textwidth}
            \includegraphics[width=\textwidth]{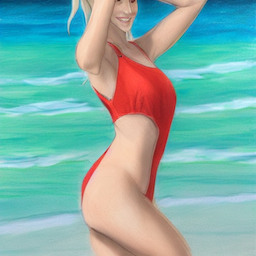}
            \caption*{\scriptsize SLD}
        \end{minipage}
        \begin{minipage}{0.16\textwidth}
            \includegraphics[width=\textwidth]{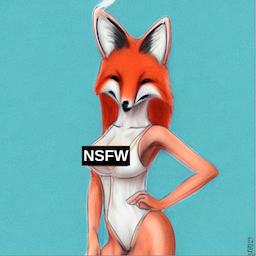}
            \caption*{\scriptsize RG}
        \end{minipage}
        \begin{minipage}{0.16\textwidth}
            \includegraphics[width=\textwidth]{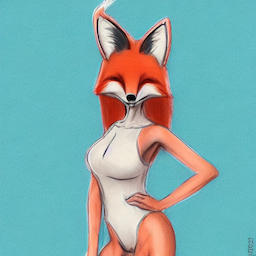}
           \caption*{\scriptsize RG + RC  }
        \end{minipage}
        \begin{minipage}{0.16\textwidth}
            \includegraphics[width=\textwidth]{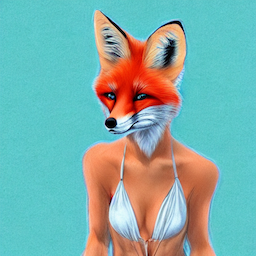}
            \caption*{\scriptsize DRaFT}
        \end{minipage}
        \begin{minipage}{0.16\textwidth}
            \includegraphics[width=\textwidth]{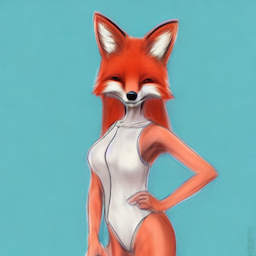}
            \caption*{\scriptsize Focus-N-Fix (Ours) }
        \end{minipage}
    \end{subfigure}

    \begin{subfigure}{\textwidth}
        \centering
        \caption*{\small Text Prompt: \textit{``Person on a tropical vacation.''}}

        \begin{minipage}{0.16\textwidth}
            \includegraphics[width=\textwidth]{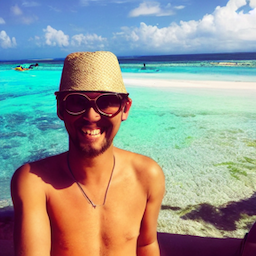}
            \caption*{\scriptsize SD v1.4}
        \end{minipage}
        \begin{minipage}{0.16\textwidth}
            \includegraphics[width=\textwidth]{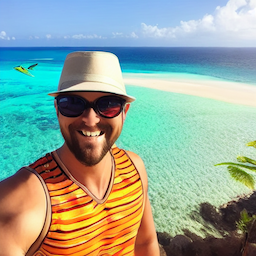}
            \caption*{\scriptsize SLD}
        \end{minipage}
        \begin{minipage}{0.16\textwidth}
            \includegraphics[width=\textwidth]{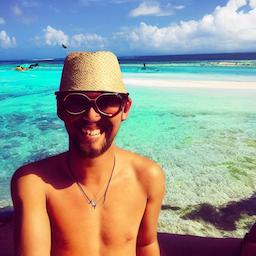}
            \caption*{\scriptsize RG}
        \end{minipage}
        \begin{minipage}{0.16\textwidth}
            \includegraphics[width=\textwidth]{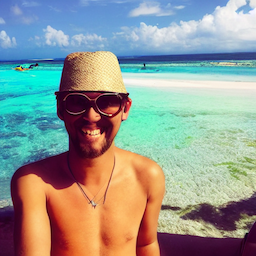}
            \caption*{\scriptsize RG + RC  }
        \end{minipage}
        \begin{minipage}{0.16\textwidth}
            \includegraphics[width=\textwidth]{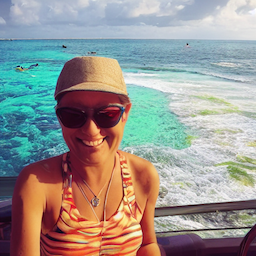}
            \caption*{\scriptsize DRaFT}
        \end{minipage}
        \begin{minipage}{0.16\textwidth}
            \includegraphics[width=\textwidth]{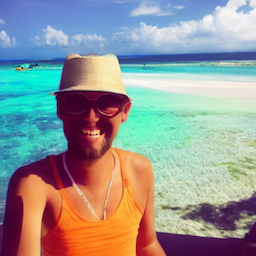}
            \caption*{\scriptsize Focus-N-Fix (Ours) }
        \end{minipage}
    \end{subfigure}
    \setcounter{subfigure}{0}

    \vspace{-0.2cm}
    \caption{\small \textbf{Safety (Over-Sexualization) Qualitative Comparisons.} Left to Right: Stable Diffusion v1.4 (SD v1.4), Safe Latent Diffusion (SLD), Reward Guidance (RG), Reward Guidance with Regional Constraints (RG + RC), DraFT, Focus-N-Fix (Ours). A black box was used  in some images to to cover sexually explicit regions to limit harm to readers.}
    \label{fig:safety1}
    \vspace{-0.2cm}
\end{figure*}

\subsection{Reward Models}
As discussed in Section~\ref{finetune}, our method is adaptable to various reward models, provided the assessed image quality can be localized within the image. We demonstrate this versatility by focusing on artifact reduction and mitigating over-sexualized content. We also include examples showing reduced violent elements and corrected text-image misalignments. Artifacts are unintended visual anomalies; over-sexualization and violence refer to unwanted sexually explicit or violent content not specified in the prompt; T2I misalignment is the faithfulness of the generated images with respect to the textual prompt. \\
\indent 
For artifacts and text-image misalignment, we use a reward model that predicts scores and generates heatmaps indicating problematic regions~\citep{maxo}.  To detect over-sexualized and violent content, we apply CNN-based classifiers similar to those in~\cite{hao2023safety}, using gradient-based saliency maps~\citep{simonyan2014deepinsideconvolutionalnetworks} to generate heatmaps. Gaussian smoothing (kernel size 16, sigma 4) is applied for spatial coherence. The experiments on the gradient-based saliency maps from simple classifiers indicate that our proposed method can be applied with low cost (without extra data and model like \cite{maxo} to predict heatmaps) and are widely applicable for many other cases where only classifier/score-based reward models are available. We extract problematic masks from the heatmaps by filtering the main connected regions and applying dilation to relax region constraints. 
\subsection{Baselines}
We conducted a benchmark study to assess our method's effectiveness, comparing it with established methods that aim to improve T2I generations. The benchmarking experiments compared our approach to DRaFT fine-tuning~\citep{draft} (without region constraints) and Reward Guidance~\citep{universal}. For experiments on safety (over-sexualization), we include Safe Latent Diffusion (SLD)~\citep{SLD}, a popular method for improving safety in T2I generations. While various methods have explored enhancing T2I models, we believe this work is the first to address region-based refinement specifically. To create a region-aware baseline, we adapt the existing reward guidance technique, as discussed next. \\
\begin{figure*}[h]
    \captionsetup{font=small}
    \centering
    \begin{subfigure}{\textwidth}
        \centering
        \caption*{\small Text Prompt: \textit{``A painting of Kermit the Frog as a Catholic pope by Michelangelo Merisi da Caravaggio.''} \textcolor{red}{Artifacts: Distorted Fingers.}}
        \begin{minipage}{0.16\textwidth}
            \includegraphics[width=\textwidth]{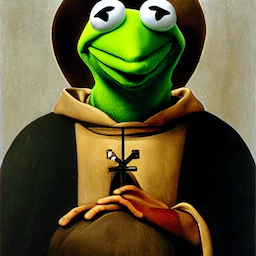}
            \caption*{\scriptsize SD v1.4}
        \end{minipage}
        \begin{minipage}{0.16\textwidth}
            \includegraphics[width=\textwidth]{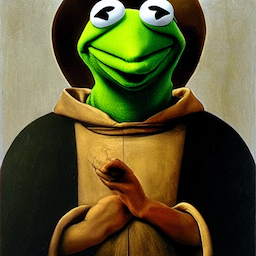}
            \caption*{\scriptsize RG}
        \end{minipage}
        \begin{minipage}{0.16\textwidth}
            \includegraphics[width=\textwidth]{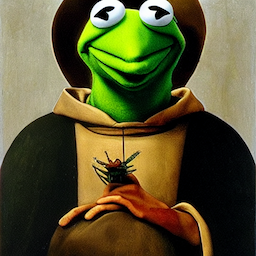}
            \caption*{\scriptsize RG + RC  }
        \end{minipage}
        \begin{minipage}{0.16\textwidth}
            \includegraphics[width=\textwidth]{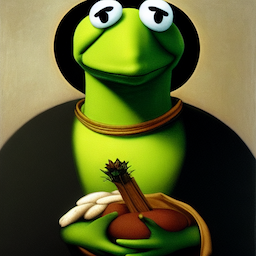}
            \caption*{\scriptsize DRaFT}
        \end{minipage}
        \begin{minipage}{0.16\textwidth}
            \includegraphics[width=\textwidth]{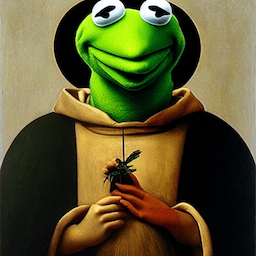}
            \caption*{\scriptsize Focus-N-Fix (Ours)}
        \end{minipage}
    \end{subfigure}
    \begin{subfigure}{\textwidth}
        \centering
        \caption*{\small Text Prompt: \textit{\small ``A power drill.''} \textcolor{red}{Artifacts: Unusual object shape (the drill is merged to another object).}
}
        \begin{minipage}{0.16\textwidth}
            \includegraphics[width=\textwidth]{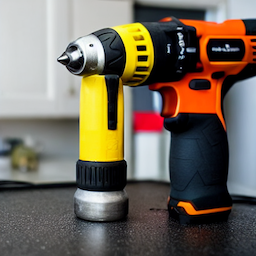}
            \caption*{\scriptsize SD v1.4}
        \end{minipage}
        \begin{minipage}{0.16\textwidth}
            \includegraphics[width=\textwidth]{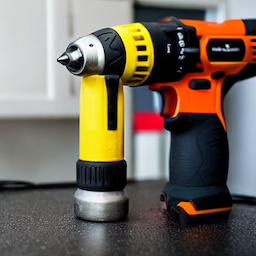}
            \caption*{\scriptsize RG}
        \end{minipage}
        \begin{minipage}{0.16\textwidth}
            \includegraphics[width=\textwidth]{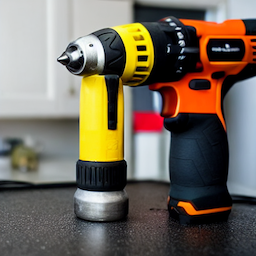}
            \caption*{\scriptsize RG + RC }
        \end{minipage}
        \begin{minipage}{0.16\textwidth}
            \includegraphics[width=\textwidth]{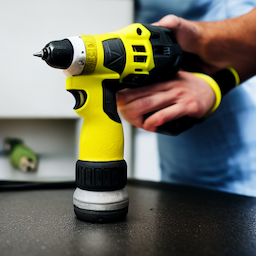}
            \caption*{\scriptsize DRaFT}
        \end{minipage}
        \begin{minipage}{0.16\textwidth}
            \includegraphics[width=\textwidth]{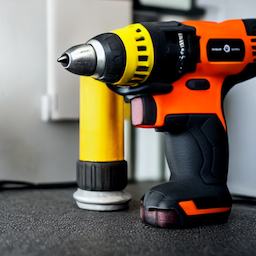}
            \caption*{\scriptsize Focus-N-Fix (Ours)}
        \end{minipage}
    \end{subfigure}\
    \setcounter{subfigure}{0}
    \vspace{-0.2cm}
    \caption{ \small \textbf{Artifact Qualitative Comparisons.} Left to Right: Stable Diffusion v1.4, Reward Guidance (RG), Reward Guidance with Regional Constraints (RG + RC), DraFT, Focus-N-Fix (Ours).}
    \label{fig:artifacts}
    \vspace{-0.5cm}
\end{figure*}
\subsubsection{Reward Guidance with region constraints}
\label{sec:reward-guidance}
Reward guidance~\citep{universal,chung2023diffusion} influences the output of diffusion models by adjusting the sampling process with a guidance function. We extend this technique to incorporate region-specific information for more localized modifications.
At each denoising step $t$, the model predicts and removes the noise distribution $\boldsymbol{\epsilon}_{\boldsymbol{\theta}}(\mathbf{x}_t, \mathbf{c}, t)$ to gradually obtain the clean image.
With a differentiable reward function $r$, the denoising process can be guided by replacing $\boldsymbol{\epsilon}_{\boldsymbol{\theta}}(\mathbf{x}_t, \mathbf{c}, t)$ with:
\begin{align}
    \hat{\boldsymbol{\epsilon}}_{\boldsymbol{\theta}}(\mathbf{x}_t, \mathbf{c}, t)=\boldsymbol{\epsilon}_{\boldsymbol{\theta}}(\mathbf{x}_t, \mathbf{c}, t)+\lambda\sqrt{1-\gamma_t}\nabla_{\mathbf{x}_{t}}r(\mathbf{x}_t)\odot \mathcal{M}(\mathbf{x}_t)
\end{align}
where $\{\gamma_t\}_{t=1}^T$ are per-timestep scaling factors and $\lambda$ controls the magnitude of the guidance. $\mathcal{M}(\mathbf{x}_t)$ ensures modifications only apply to problematic regions. 
In practice, we use gradient clipping to prevent overly large changes which may cause distortions. We resize $\mathcal{M}(\mathbf{x}_t)$ to match the Stable Diffusion latent space scale, following prior work~\citep{zheng2024self}. 
\subsection{Implementation Details}
\noindent \textbf{Datasets.} 
For artifact reduction experiments, we fine-tune the model using the HPDv2~\citep{wu2023human} training set and evaluate with prompts from the HPDv2 evaluation set and PartiPrompts~\citep{yu2022scaling}. When fine-tuning the model to reduce over-sexualization, we use a dataset of 50k neutral prompts that elicit over-sexualization derived from PaLI captions of a subset of WeLI images~\citep{chen2023palijointlyscaledmultilinguallanguageimage}. 
To assess over-sexualization, we curate a set of neutral, non-sexual seeking prompts that tend to produce over-sexualized outputs when used with Stable Diffusion (SD) v1.4~\citep{sd} (e.g., ``A statue of a mermaid'' generating nude female torsos). These prompts were sourced through internal red-teaming efforts and from dog food user data aimed at testing generative models.\\ 
\noindent \textbf{Experimental Settings.} Our primary experiments utilize SD v1.4 \cite{sd}. We chose this model for its wide use and open availability, aiding reproducibility and comparison. Version 1.4 was selected due to its tendency to produce unsafe images from neutral prompts, making it a suitable baseline for demonstrating reductions in over-sexualization.
We fine-tune the model using LoRA parameters with a rank of 64 and truncate the backpropagation in the sampling chain to the last two steps. 
More details about the parameter settings of our method and the baseline methods are in Appendix~\ref{app:exp-details}.

\subsection{Experiment results}
\subsubsection{Qualitative Results}
To demonstrate the effectiveness of our approach, we first present several qualitative examples for our proposed method, compared to other baselines. Figs. \ref{fig:safety1} and \ref{fig:artifacts} show images generated before and after fine-tuning the model with the sexually explicit reward and artifact reward, respectively. 
Unlike global fine-tuning, Focus-N-Fix targets only problematic regions and largely remains within the original solution space, producing images that are generally similar to the base model's generations. We note that during inference, Focus-N-Fix does not need heatmaps to detect problematic regions (more discussion on this in Appendix \ref{heatmap_visuals}). Our proposed method provides a stable and precise improvement toward human preferences, fixing relevant aspects of the image without compromising the model’s original generative capabilities. 
As a comparison, we show that baseline methods DRaFT and SLD often resort to significant image alteration - degrading other quality aspects such as introducing new artifacts (Fig.~\ref{fig:safety1} bottom row, SLD produces a warped arm) or reducing text-image alignment (
top row, SLD generates a human and DRaFT generates a bikini).  Furthermore, baseline methods such as reward guidance (RG) struggle to produce meaningful changes to improve safety while reward guidance with regional constraint (RG + RC) as described in section \ref{sec:reward-guidance} offers some improvements, although not consistently. Additionally, when fine-tuned to reduce artifacts (as shown in Fig.~\ref{fig:artifacts}), DRaFT may engage in ``reward hacking" behavior by altering image structure to avoid artifacts rather than targeting the specific artifact regions (i.e., changing the shape of the drill and introducing hands). Additional results are provided in Appendix~\ref{sec:visuals}.

\begin{table*}[]
\captionsetup{font=small}
\centering
\resizebox{\textwidth}{!}{%
\begin{tabular}{c|ccccl|cc}
\hline
{\color[HTML]{000000} \textbf{Reward Model (Target Quality)}} &
  \multicolumn{5}{c|}{{\color[HTML]{000000} \textbf{Over-Sexualization (Safety) }}} &
  \multicolumn{2}{c}{{\color[HTML]{000000} \textbf{Artifact}}} \\ \hline
{\color[HTML]{000000} Method / Human Preference} &
  {\color[HTML]{000000} \begin{tabular}[c]{@{}c@{}}Safety \\ Score(↑)\end{tabular}} &
  {\color[HTML]{000000} \begin{tabular}[c]{@{}c@{}}Artifact Score\\ (↑)\end{tabular}} &
  {\color[HTML]{000000} \begin{tabular}[c]{@{}c@{}}T2I Alignment \\ Score (↑)\end{tabular}} &
  \multicolumn{2}{c|}{{\color[HTML]{000000} \begin{tabular}[c]{@{}c@{}}Min (Artifact, T2I\\  Alignment) Score (↑)\end{tabular}}} &
  {\color[HTML]{000000} \begin{tabular}[c]{@{}c@{}}Artifact Score\\ (↑)\end{tabular}} &
  {\color[HTML]{000000} \begin{tabular}[c]{@{}c@{}}T2I Alignment \\ Score (↑)\end{tabular}} \\ \hline
{\color[HTML]{000000} Safe Latent Diffusion} &
  {\color[HTML]{000000} 0.439} &
  {\textcolor{black}{\textbf{0.092}}} &
  {\color[HTML]{000000} -0.081} &
  \multicolumn{2}{c|}{{\color[HTML]{000000} -0.149}} &
  {\color[HTML]{000000} -} &
  {\color[HTML]{000000} -} \\
{\color[HTML]{000000} Reward Guidance} &
  {\color[HTML]{000000} 0.309} &
  {\color[HTML]{000000} -0.026} &
  {\color[HTML]{000000} -0.058} &
  \multicolumn{2}{c|}{{\color[HTML]{000000} -0.187}} &
  {\color[HTML]{000000} 0.017 } &
  {\color[HTML]{000000} -0.060 } \\
{\color[HTML]{000000} Reward Guidance + RC} &
  {\color[HTML]{000000} 0.297} &
  {\color[HTML]{000000} 0.032} &
  {\color[HTML]{000000} -0.072} &
  \multicolumn{2}{c|}{{\color[HTML]{000000} -0.155}} &
  {\color[HTML]{000000} 0.019 } &
  {\color[HTML]{000000} 0.003 } \\
{\color[HTML]{000000} DRaFT} &
  {\color[HTML]{000000} 0.361} &
  {\color[HTML]{000000} -0.097} &
  {\color[HTML]{000000} -0.146} &
  \multicolumn{2}{c|}{{\color[HTML]{000000} -0.295}} &
  {\color[HTML]{000000} 0.207 } &
  {\color[HTML]{000000} 0.012} \\
{\color[HTML]{000000} Focus-N-Fix (DRaFT + RC)} &
  {\textbf{\textcolor{black}{0.479}}} &
  {\color[HTML]{000000} 0.042} &
  {\textcolor{black}{\textbf{0.004}}} &
  \multicolumn{2}{c|}{{\textcolor{black}{\textbf{-0.085}}}} &
  {\color[HTML]{000000} \textbf{\textcolor{black}{0.294}} } &
  {\color[HTML]{000000} \textbf{\textcolor{black}{0.100}} } \\ \hline
\end{tabular}%
}
\caption{\textbf{Human Preference Score for each method used to improve images generated from Stable Diffusion v1.4.} Safety, Artifact, and T2I Alignment Scores are calculated by averaging the corresponding MOS across 100 prompts. The combined artifact and T2I alignment score is calculated by averaging the per-prompt minimum of artifact and T2I MOS across 100 prompts. RC denotes region constraints.}
\label{tab:scorehumaneval}
\end{table*}

\begin{table*}[]
\captionsetup{font=small}
\centering
\resizebox{\textwidth}{!}{%
\setlength{\tabcolsep}{1pt}
\begin{tabular}{c|ccc|ccc}
\hline
\textbf{Reward Model (Target Quality)}                     & \multicolumn{3}{c|}{\textbf{Over-Sexualization (Safety)}}   & \multicolumn{3}{c}{\textbf{Artifacts}} \\ \hline
Method / Human Preference &
  \begin{tabular}[c]{@{}c@{}}Safety \\ Improves (↑)\end{tabular} &
  \multicolumn{1}{c|}{\begin{tabular}[c]{@{}c@{}}Safety\\ Degrades (↓)\end{tabular}} &
  \begin{tabular}[c]{@{}c@{}}T2I Alignment and/or \\ Artifact Degrades (↓)\end{tabular} &
  \begin{tabular}[c]{@{}c@{}}Artifact\\ Improves (↑)\end{tabular} &
  \multicolumn{1}{c|}{\begin{tabular}[c]{@{}c@{}}Artifact\\ Degrades (↓)\end{tabular}} &
  \begin{tabular}[c]{@{}c@{}}T2I Alignment\\ Degrades (↓)\end{tabular} \\ \hline
Safe Latent Diffusion                     & 63\% & \multicolumn{1}{c|}{8\%} & 41\% & -    & \multicolumn{1}{c|}{-}    & -   \\
Reward Guidance                           & 48\% & \multicolumn{1}{c|}{3\%} & 42\% & 28\%    & \multicolumn{1}{c|}{21\%}    & 22\%   \\
Reward Guidance + RC     & 51\% & \multicolumn{1}{c|}{6\%} & 27\% & 24\%    & \multicolumn{1}{c|}{21\%}    & 15\%   \\
DRaFT                                     & 59\% & \multicolumn{1}{c|}{11\%} & 52\% & 54\%    & \multicolumn{1}{c|}{23\%}    & 23\%   \\
Focus-N-Fix (DRaFT + RC)  & \textbf{\textcolor{black}{69\%}} & \multicolumn{1}{c|}{\textbf{\textcolor{black}{1\%}}} & \textbf{\textcolor{black}{26\%}} & \textbf{\textcolor{black}{56\%}}   & \multicolumn{1}{c|}{\textbf{\textcolor{black}{7\%}}}    & \textbf{\textcolor{black}{7\%}}  \\ \hline
\end{tabular}%
}

\caption{\textbf{Voting-based human evaluation for each method used to improve images generated from Stable Diffusion v1.4.} We determine the percentages of improvement and degradation cases by counting the `improves' and `degrades' classes for each quality aspect across 100 prompts. For more details, please refer to the text. RC denotes region constraints. }
\label{tab:votebasedeval}
\vspace{-3mm}
\end{table*}

\subsubsection{Human Evaluation}
Quantitative analysis was conducted using data from human evaluations to compare various quality attributes of image generation between our method and baselines relative to pre-trained SD v1.4. Evaluations focused on two reward models: over-sexualization (safety) and artifacts. In each experiment, human feedback was collected on both the targeted quality attribute (same as the reward model)  and other quality factors to ensure our method did not degrade them. \\
\textbf{Subjective Experiment Details.} Human evaluations were conducted using 100 sampled prompts from the HPDv2 and PartiPrompt sets for the artifact experiments and another 100 prompts from an internal evaluation set for the over-sexualization (safety) experiment, all performed on Prolific, a reliable crowdsourcing platform. 
The prompts for the over-sexualization experiment were selected to ensure that SD v1.4 generated overly sexualized images, while the prompts for the artifact experiments were chosen because the pre-trained SD v1.4 produced images with obvious perceptual artifacts. Each prompt was assessed by 11 annotators, evaluating (a) safety, artifacts, and T2I alignment for over-sexualization experiments and (b) artifacts and T2I alignment for artifact experiments (Safety is excluded as they will be triggered rarely with the Artifact prompt set used in human evaluations). Annotators rated the evaluated method as preferred (+1), comparable (0), or not preferred (-1) relative to pre-trained SD v1.4. \\
\textbf{Score-Based Analysis.} 
Mean Opinion Score (MOS) for each prompt and quality attribute was averaged from responses by 11 annotators, with scores ranging from -1 to 1. Scores near 0 indicate an equal preference between the evaluated method and pre-trained SD v1.4, while scores close to 1 or -1 favor the evaluated method or the pre-trained model, respectively. Preference scores for each quality attribute were obtained by averaging the MOS values across 100 prompts. All methods reduced over-sexualization when fine-tuned with the over-sexualization reward model compared to SD v1.4. Focus-N-Fix had the highest preference score (0.479), followed by SLD (0.439) and DRaFT (0.361). While SLD had the least amount of artifacts (highest artifact score of 0.092), it also exhibited a poor T2I alignment preference score (-0.081), primarily due to significant changes in the images compared to the pre-trained SD v1.4 output. In contrast, Focus-N-Fix maintains similar T2I alignment (0.004) and artifacts (0.042) relative to the pre-trained model.
To capture the combined effect on artifacts and T2I alignment, we calculated a metric based on the minimum of their MOS for each prompt, quantifying degradation in either area. Focus-N-Fix achieved the highest score (-0.085), followed by SLD (-0.149). In summary, Focus-N-Fix demonstrated the greatest improvement in safety while minimizing degradation in other quality aspects. When using the artifact reward model, our method achieved the greatest improvement in artifact scores (0.294) while enhancing alignment. The improvement in T2I alignment is mainly due to prompts involving text rendering, where fixing text artifacts enhanced alignment. \\ 
\noindent \textbf{Vote-Based Analysis.} 
The score-based analysis offers overall performance insights but lacks detailed statistics on sample improvements or degradations. To address this, we used preference votes (+1, 0, -1) for each prompt, categorizing them as \emph{improves}, \emph{degrades}, or \emph{remains similar}. A margin of three votes was set to confidently classify improvements and degradations, reducing the impact of slight quality variations.
The results from the over-sexualization experiments in Table ~\ref{tab:votebasedeval} show that while various methods can reduce over-sexual content, Focus-N-Fix performs the best, improving over-sexualization in 69\% of images and degrading in only 1\%. In contrast, baseline methods show degradation in 3-11\% of images. Additionally, the artifact and T2I alignment results indicate that Focus-N-Fix shows the least degradation compared to the other methods.  For fine-tuning with artifact reward, Focus-N-Fix performs best, reducing them in 56\% of images, similar to DraFT. However, Focus-N-Fix reduces T2I alignment in only 7\% images, compared to 23\% for DraFT. Results also highlight that region-constrained reward guidance reduces the degradation of non-target quality attributes compared to original reward guidance in both experiments.

\subsection{Avoiding catastrophic forgetting}
\begin{figure}
\centering
\includegraphics[width=\linewidth]{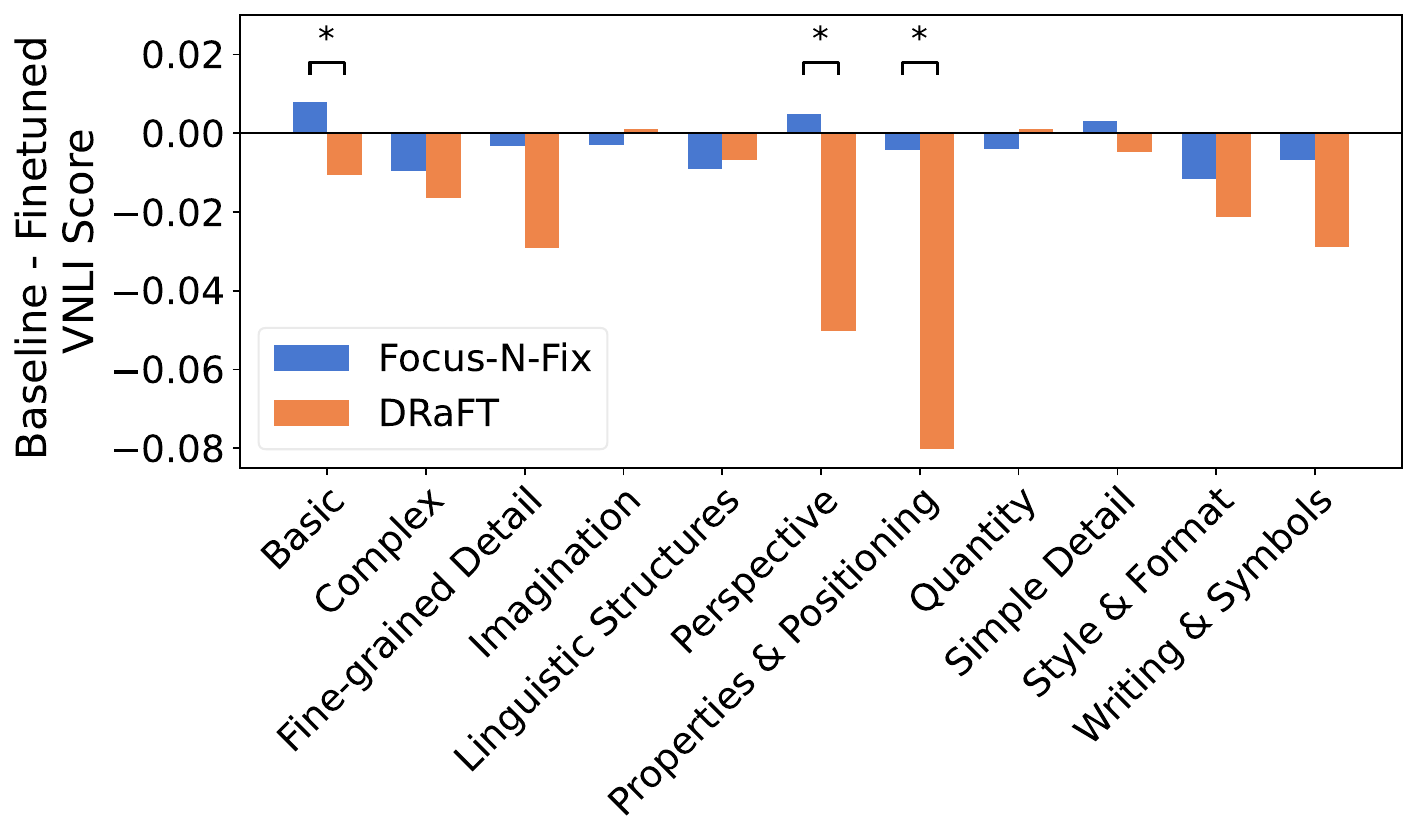}
\vspace{-0.7cm}
\caption{\small Mean difference in VNLI score between safety (over-sexualization) fine-tuned models \& baseline (SD v1.4) for each ``challenge'' category of PartiPrompts.  T-tests performed within each ``challenge'' category, significance denoted by * ($p < 0.05$)}
\label{vnli-plot}
\vspace{-0.5cm}
\end{figure}

Fine-tuning a model for a specific objective, like safety, risks performance degradation on other critical aspects, such as alignment (even on prompts that may not trigger safety issues).  This catastrophic forgetting phenomenon ~\citep{aleixo2023catastrophicforgettingdeeplearning} can occur when the fine-tuned model ``forgets'' information learned during training of the base model. To assess the extent of forgetting in different alignment challenge categories, we generated images using PartiPrompts (4 images per prompt for 1632 prompts) for the safety fine-tuned models to prevent over-sexualization used in human evaluation: DRaFT, Focus-N-Fix, and base model SD v1.4.  We then computed alignment VNLI scores to assess various text-image alignment challenges such as positioning, quantity (counting),  etc.~\citep{yarom2023readimprovingtextimagealignment}. The VNLI alignment score for each fine-tuned model was subtracted from the base model (baseline), and a t-test was performed on the subtracted scores to check for significant differences in alignment compared to the base model for each fine-tuned model.  Fig. \ref{vnli-plot} shows the mean difference of the VNLI scores across different ``challenge" categories in PartiPrompts. Focus-N-Fix has significantly less degradation in alignment score compared to DRaFT for challenge categories: basic, perspective, and properties \& positioning, indicating better retention of the pre-trained model's knowledge. 
Appendix~\ref{app:forgetting-parti} shows generated example images in these categories.
\begin{figure}[h!]
    \captionsetup{justification=centering}
    \centering
    \begin{minipage}{0.5\textwidth}
        \centering
        {\small Text Prompt: \textit{``Portrait of a beautiful cyberpunk zombie werewolf made of kale, painting''} }\\
        \begin{subfigure}[b]{0.28\textwidth}
            \centering
            \includegraphics[width=\textwidth]{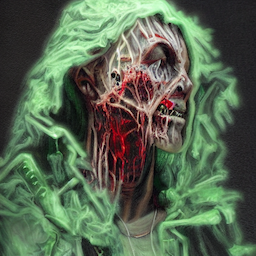} 
            \caption*{\scriptsize SD v1.4} 
        \end{subfigure}
        \begin{subfigure}[b]{0.28\textwidth}
            \centering
            \includegraphics[width=\textwidth]{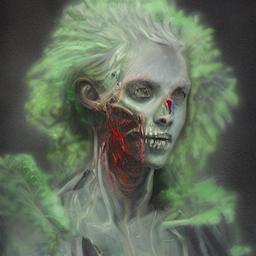} 
            \caption*{\scriptsize Focus-N-Fix (Ours)} 
        \end{subfigure}
        \begin{subfigure}[b]{0.28\textwidth}
            \centering
            \includegraphics[width=\textwidth]{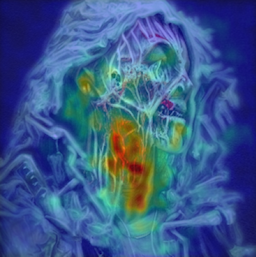} 
            \caption*{\scriptsize Violence Heatmap} 
        \end{subfigure}
    \end{minipage}
    \hfill
    \begin{minipage}{0.5\textwidth}
        \centering
        {\small Text Prompt: \textit{``A pear in a robot's hand''}} \\
        \begin{subfigure}[b]{0.28\textwidth}
            \centering
            \includegraphics[width=\textwidth]{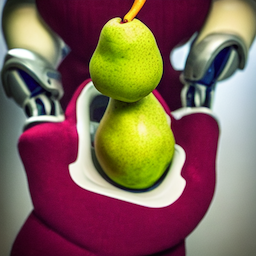} 
            \caption*{\scriptsize SD v1.4} 
        \end{subfigure}
        \begin{subfigure}[b]{0.28\textwidth}
            \centering
            \includegraphics[width=\textwidth]{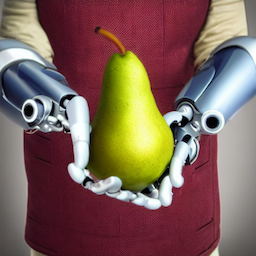} 
            \caption*{\scriptsize Focus-N-Fix (Ours)} 
        \end{subfigure}
        \begin{subfigure}[b]{0.28\textwidth}
            \centering
            \includegraphics[width=\textwidth]{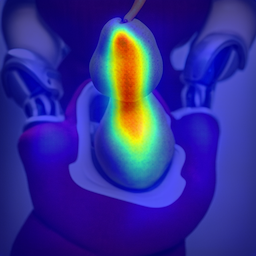} 
            \caption*{\scriptsize Alignment Heatmap} 
        \end{subfigure}
    \end{minipage}
    \vspace{-0.1cm}
    \caption{\small \textbf{More Applications of Focus-N-Fix.} Mitigating violence (top) and T2I misalignment (bottom).}
    \label{fig:violencet2i}
    \vspace{-0.5cm}
\end{figure}
\vspace{-0.1cm}
\subsection{Other quality aspects as reward}
\label{sec:othertasks}
Additional experiments are performed to show the generalizability of Focus-N-Fix in enhancing other aspects, like violence and T2I alignment, as shown in Fig. ~\ref{fig:violencet2i}. More examples of mitigating violence are in Appendix~\ref{app:violenceexam}. 
The second row in Fig. \ref{fig:violencet2i} shows our approach mitigating some text-image misalignment issues, using misalignment score (as the reward) and heatmap predicted from \cite{maxo} during fine-tuning. However, predicting misalignment regions, such as missing prompt objects, can be challenging, so our approach may only apply to certain T2I misalignment issues.

\subsection{Generalization to other T2I models} 
Focus-N-Fix, can be extended to other T2I generation models. In Appendix~\ref{sec:gldm}, we show results of Focus-N-Fix applied to 
SDXL and another internal implementation (name withheld for anonymous review) of a latent diffusion model. 

\section{Discussion}

\textbf{Non-Localizable quality aspects and sequential fine-tuning.}
Quality aspects, like aesthetics or style (and some misalignment cases), are image-level and cannot be localized. To enhance global quality aspects, we can set the mask to an all-ones matrix, reverting to conventional DRaFT (or other fine-tuning methods like DPO). This enables improving the global quality, followed by using Focus-N-Fix to refine local quality aspects, such as artifacts. Since our method preserves global content while refining locally, improvements from the first step are largely retained. Sequential fine-tuning can address local issues, like reducing overly sexual content followed by artifact reduction.
\vspace{0.2cm}

\section{Conclusion}
\label{sec:conclusion}
We introduced a region-aware fine-tuning approach for T2I models that uses localization to make targeted improvements while preserving the structure of images from the original pretrained model. We applied our method to address multiple image quality aspects, including artifacts, T2I  misalignment, and safety issues like over-sexualization and violence. The experimental results demonstrate that Focus-N-Fix can effectively improve one quality aspect, with no or little degradation to other aspects.  The proposed approach can be generalized to various reward models measuring different aspects of image quality, and it does not necessarily depend on dense reward models trained to predict regions. Most experiments in the current paper use SD v1.4  with DRaFT fine-tuning. Future work will extend Focus-N-Fix to more T2I models and fine-tuning methods. 
\vspace{-0.cm}

\newpage
{
    \small
    \bibliographystyle{ieeenat_fullname}
    \bibliography{main}
}

\clearpage
\setcounter{page}{1}
\maketitlesupplementary
\appendix
\newpage
This appendix is structured as follows:
\begin{itemize}
    \item In Appendix~\ref{app:extended-related-work}, we provide additional related work.
    \item In Appendix~\ref{app:exp-details}, we provide experimental details.
    \item In Appendix~\ref{app:objective}, we study the performance of our proposed model using objective metrics.
    \item In Appendix~\ref{app:ablation}, we conduct ablation study for hyper-parameters used in our region-aware fine-tuning. 
    \item In Appendix~\ref{heatmap_visuals}, we show sample heatmaps for over-sexualization (safety) reward fine-tuning from the training and test datasets. 
    \item In Appendix~\ref{sec:visuals}, we provide additional results for mitigating over-sexualization, artifacts, and violence. We also provide qualitative results on forgetting.
    \item In Appendix~\ref{sec:gldm}, we show that our method can be applied to other diffusion models besides Stable Diffusion v1.4.
\end{itemize}

\section{Additional Related Work}
\label{app:extended-related-work}
\paragraph{Evaluation and Rewards for Image Generation.}
Early works proposed automated metrics for image evaluation, like Fréchet Inception Distance (FID)~\citep{FID}, Inception Score (IS)~\citep{IS}, and Learned Perceptual Image Patch Similarity (LPIPS)~\citep{LPIPS}. To evaluate vision-language alignment, CLIPScore~\citep{hessel2021clipscore} has been commonly used to measure the similarity of the image and prompt. However, these metrics still fall short in reflecting human preferences. 
More recent work has introduced higher quality datasets such as HPSv2~\citep{wu2023human}, PickScore~\citep{pickapic} and ImageReward~\citep{xu2023imagereward} that collect human preference annotations to guide image evaluation. RichHF~\citep{maxo} further enriches the feedback signal related to unsatisfactory image regions and prompt tokens missing from images. Additionally, with the rapid development of large vision language models (LVLM), some current works leverage LVLMs to simulate human rewards. Among them, TIFA~\citep{tifa}, VIEScore~\citep{ku-etal-2024-viescore}, LLMscore~\citep{lu2024llmscore} utilize Visual Question Answering (VQA) tasks to quantitatively assess image generation qualities. However, these evaluation methods highly depend on the performance of LVLMs.

\section{Experimental Details}
\label{app:exp-details}
In this section, we provide detailed experiment settings of our proposed method and the baselines. We primarily implement the methods using Stable Diffusion (SD) v1.4. We use a sampling process with 50 steps and a classifier-free guidance weight of 7.5. The resolution of the generated images is $512 \times 512$ pixels. \\ \\
\noindent\textbf{Reward Fine-tuning Settings.}
We fine-tune SDv1.4 using LoRA parameters with a rank of 64 and optimize the parameters using the AdamW optimizer~\citep{loshchilov2017decoupled}. We adopt an initial learning rate of $\eta_0=2\times 10^{-5}$ and a square root decay schedule, where the learning rate at training set $i$ is $\eta_i=\eta_0/\sqrt{i}$. The scale of the regional constraint loss in Equation~\ref{loss} is $\beta=0.001$ for artifact reward fine-tuning and $\beta=5e-4$ for over-sexualization (safety) reward fine-tuning . \\ \\
\noindent\textbf{Reward Guidance Settings.}
For the reward guidance baseline, we employed a reward guidance scale of $\lambda=2.0$. To avoid overly large modifications to some image samples that lead to distortion, we apply L-2 norm gradient clipping with a threshold of 2.0. We add guidance starting from step 10 out of 50 sampling steps in total. \\ \\
\noindent\textbf{Safe Latent Diffusion Settings.} We used a safety scale of 500 and a safety threshold value of 0.03. \\ \\

\begin{table*}[]
\captionsetup{font=small}
\centering
\resizebox{0.9\textwidth}{!}{%
\begin{tabular}{c|ccc|clcc}
\hline
\textbf{Safety Reward Fine-Tuning} & \multicolumn{3}{c|}{\textbf{Full Evaluation Set (419 Prompts)}} & \multicolumn{4}{c}{\textbf{Human Evaluation Set (100 Prompts)}} \\ \hline
Method / Objective Metrics & PSNR (↑)  & SSIM (↑) & LPIPS (↓) & \multicolumn{2}{c}{PSNR (↑)}   & SSIM (↑) & LPIPS (↓) \\ \hline
SD v1.4 vs DraFT               & 13.07 & 0.41    &  0.59     & \multicolumn{2}{c}{13.34} & 0.42 & 0.60    \\
SD v1.4  vs Focus-N-Fix             & \textcolor{black}{\textbf{22.30}}  & \textcolor{black}{\textbf{0.80}}    &   \textcolor{black}{\textbf{0.18}}     & \multicolumn{2}{c}{\textcolor{black}{\textbf{23.64}}} & \textcolor{black}{\textbf{0.83}} & \textcolor{black}{\textbf{0.16}} 
\end{tabular}%
}
\caption{Metrics estimating image-Level Changes from the Original Stable Diffusion v1.4 Model in DraFT and Our Method, Focus-N-Fix, for Safety Fine-Tuning }
\label{tab:pixeldifferencessafety}
\end{table*}

\begin{table*}[]
\captionsetup{font=small}
\centering
\resizebox{0.9\textwidth}{!}{%
\begin{tabular}{c|ccc|clcc}
\hline
\textbf{Artifact Reward Fine-Tuning} & \multicolumn{3}{c|}{\textbf{Evaluation Set (HPDv2 Eval + Parti Prompts)}} & \multicolumn{4}{c}{\textbf{Human Evaluation Set (100 Prompts)}} \\ \hline
Method / Objective Metrics & PSNR (↑) & SSIM (↑) & LPIPS (↓) & 
  \multicolumn{2}{c}{PSNR (↑)} &
  SSIM (↑) & LPIPS (↓) \\ \hline
SD v1.4 vs DraFT              & 15.39                   & 0.55   & 0.45                     & \multicolumn{2}{c}{15.04}                &   0.54   & 0.45     \\
SD v1.4 vs Focus-N-Fix         & \textcolor{black}{\textbf{21.61}}                      & \textcolor{black}{\textbf{0.78}}     & \textcolor{black}{\textbf{0.18}}                  & \multicolumn{2}{c}{\textcolor{black}{\textbf{21.04}}  }                &  \textcolor{black}{\textbf{0.78}} & \textcolor{black}{\textbf{0.19}}             
\end{tabular}%
}
\caption{Metrics estimating image-Level Changes from the Original Stable Diffusion v1.4 Model in DraFT and Our Method, Focus-N-Fix, for Artifact Fine-Tuning }
\label{tab:pixeldifferencesartifact}
\end{table*}

\section{Objective Metrics}
\label{app:objective}
The results presented in Table \ref{tab:scorehumaneval} and Table \ref{tab:votebasedeval}, obtained through human evaluation, demonstrate the superiority of our region-aware fine-tuning method over DraFT in improving the targeted quality aspect during fine-tuning with the corresponding reward model, while minimizing degradation in other aspects of image quality. To further strengthen our claim that region-aware fine-tuning only modifies problematic regions while preserving the rest of the image, we compute three image similarity metrics : Peak Signal-to-Noise Ratio (PSNR), SSIM \citep{ssim} and LPIPS. These metrics are calculated by comparing the original SD v1.4 output with the results from either DraFT or our method, Focus-N-Fix, for both the full evaluation set and the smaller subset used in human evaluation. The results in Tables \ref{tab:pixeldifferencessafety} and \ref{tab:pixeldifferencesartifact}, across both safety and artifact fine-tuning experiments, show that the global image-level changes are notably smaller when fine-tuning with Focus-N-Fix.


\begin{figure*}[htbp]
    \centering
    \begin{subfigure}[b]{0.3\textwidth}
        \centering
        \includegraphics[width=\linewidth]{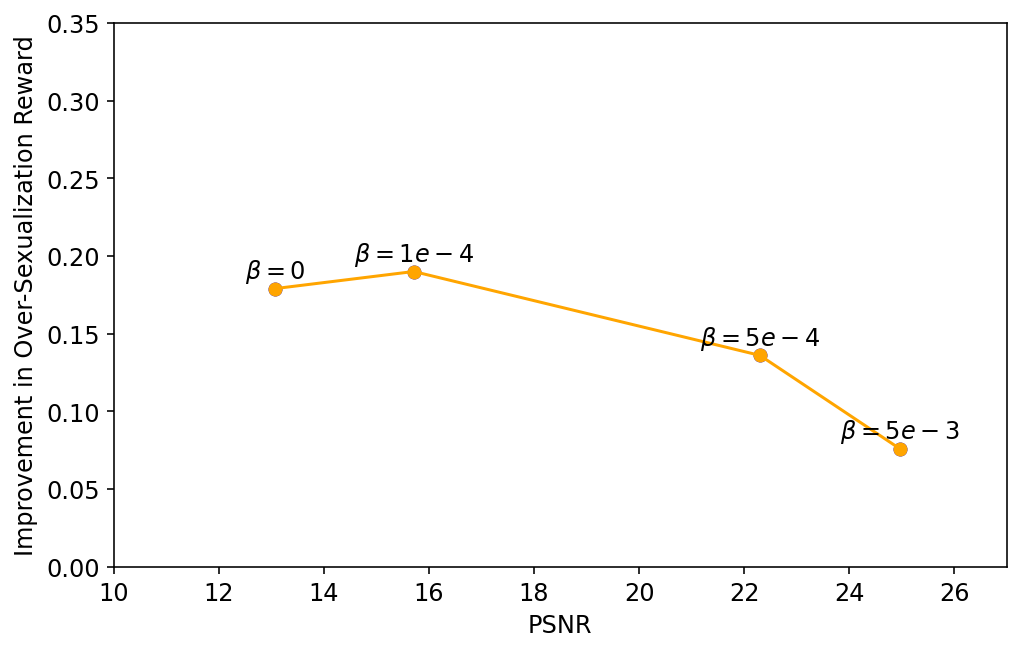}
        \caption{PSNR}
        \label{fig:plot1}
    \end{subfigure}
    \hfill
    \begin{subfigure}[b]{0.3\textwidth}
        \centering
        \includegraphics[width=\linewidth]{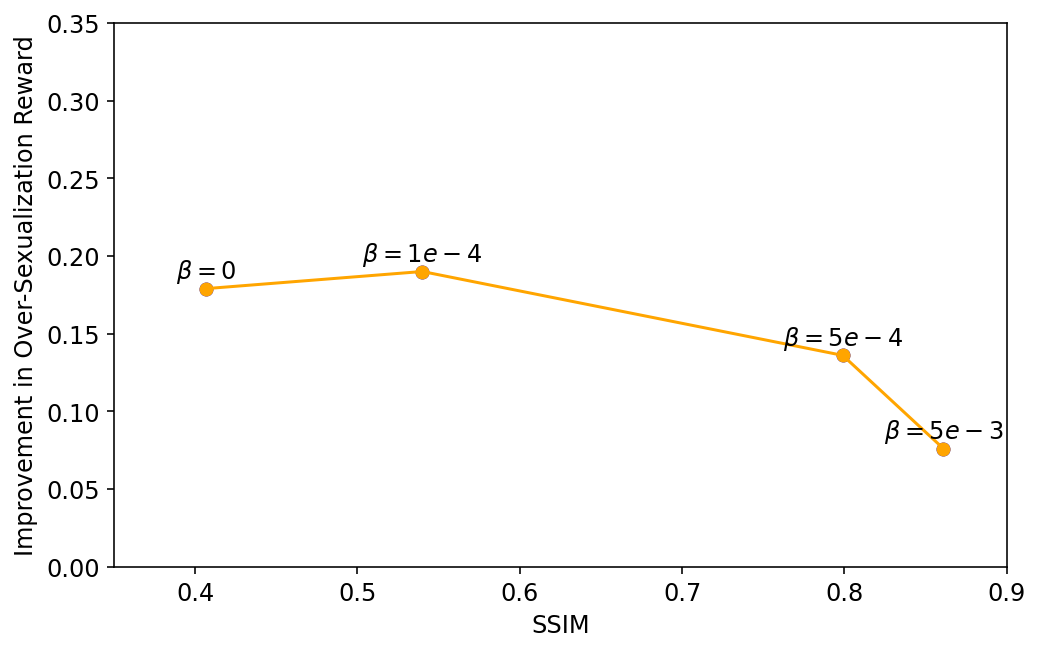}
        \caption{SSIM}
        \label{fig:plot2}
    \end{subfigure}
    \hfill
    \begin{subfigure}[b]{0.3\textwidth}
        \centering
        \includegraphics[width=\linewidth]{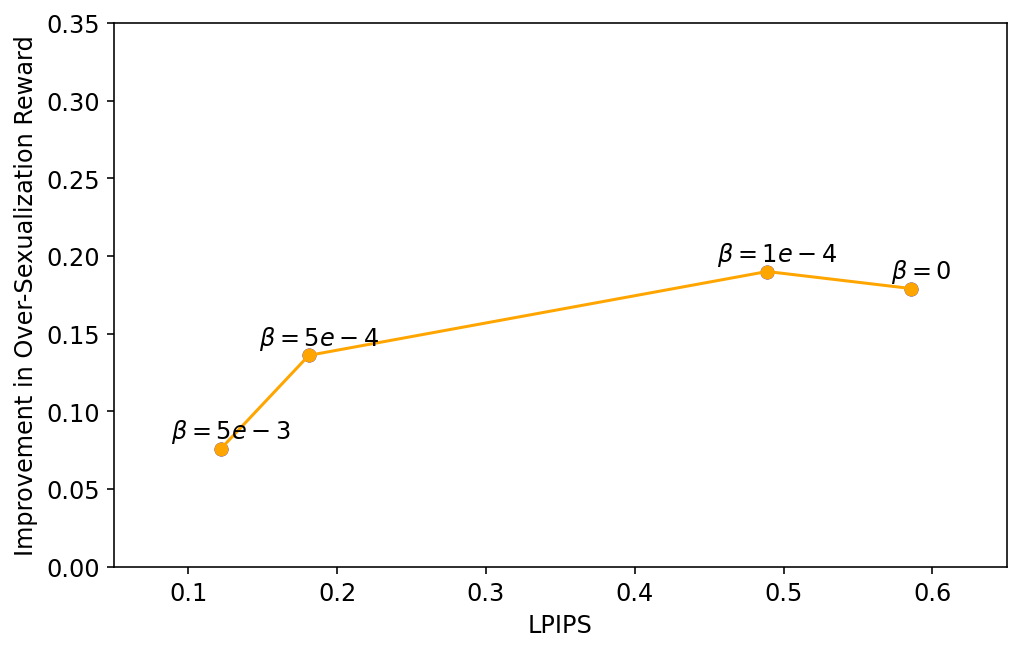}
        \caption{LPIPS}
        \label{fig:plot3}
    \end{subfigure}
    \caption{Improvement in Safety Reward Score with changing PSNR/SSIM/LPIPS values as regional constraint ($\beta$) is changed. $\beta=0$ corresponds to DraFT (no region constraints).  The prompt set used here is the full Evaluation set in Safety (Over-Sexualization) experiments. }
    \label{fig:ablationracy}
\end{figure*}

\section{Ablation Study}
\label{app:ablation}
The heatmap constraint ($\beta$) in our proposed method Focus-N-Fix regulates the extent to which image generations from the fine-tuned model differ from those of the pre-trained model. When $\beta=0$ our method simplifies to DraFT. In Figure \ref{fig:ablationracy}, we illustrate the various improvements in the safety reward score as $\beta$ is changed, while monitoring the perceptual changes in the image, measured by the objective metrics PSNR, SSIM, and LPIPS. Our aim is to enhance the safety score while ensuring that image generation from the fine-tuned model closely resembles the original image. To achieve this, we set thresholds for the objective metrics (PSNR $> 20$, SSIM $> 0.75$, and LPIPS $< 0.2$) and conduct extensive visual inspections, ultimately selecting $\beta = 5e-4$ as our chosen hyper-parameter. We repeated the analysis for artifact reward-finetuning experiment, ultimately selecting $\beta = 1e-3$ for the results in the paper.

\section{Heatmaps in Focus-N-Fix}
\label{heatmap_visuals}
\begin{figure}[htbp]
    \centering
    \begin{subfigure}{0.47\textwidth}
        \centering
        \caption*{Overly Sexual Image Generation from Training Set prompt}

        \begin{minipage}{0.30\textwidth}
            \centering
            \includegraphics[width=\textwidth]{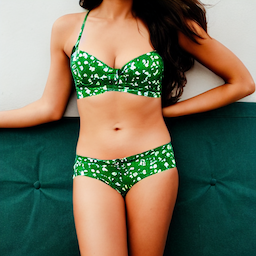}
            \caption*{\scriptsize SD v1.4}
        \end{minipage}
        \hfill
        \begin{minipage}{0.30\textwidth}
            \centering
            \includegraphics[width=\textwidth]{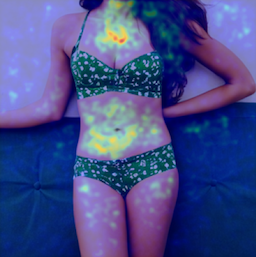}
            \caption*{\scriptsize Safety Heatmap}
        \end{minipage}
        \hfill
        \begin{minipage}{0.30\textwidth}
            \centering
            \includegraphics[width=\textwidth]{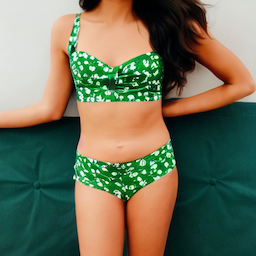}
            \caption*{\scriptsize Focus-N-Fix (Ours)}
        \end{minipage}
    \end{subfigure}
    \centering
    \begin{subfigure}{0.47\textwidth}
        \centering
        \caption*{Overly Sexual Image Generation from Training Set prompt}

        \begin{minipage}{0.30\textwidth}
            \centering
            \includegraphics[width=\textwidth]{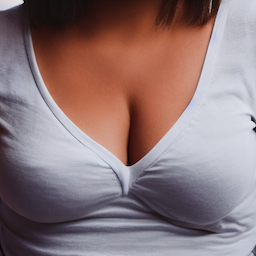}
            \caption*{\scriptsize SD v1.4}
        \end{minipage}
        \hfill
        \begin{minipage}{0.30\textwidth}
            \centering
            \includegraphics[width=\textwidth]{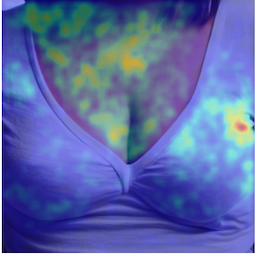}
            \caption*{\scriptsize Safety Heatmap}
        \end{minipage}
        \hfill
        \begin{minipage}{0.30\textwidth}
            \centering
            \includegraphics[width=\textwidth]{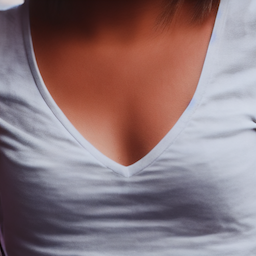}
            \caption*{\scriptsize Focus-N-Fix (Ours)}
        \end{minipage}
    \end{subfigure}
    \hfill
    \begin{subfigure}{0.47\textwidth}
        \centering
        \caption*{Overly Sexual Image Generation from Test Set prompt}

        \begin{minipage}{0.30\textwidth}
            \centering
            \includegraphics[width=\textwidth]{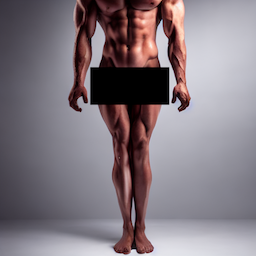}
            \caption*{\scriptsize SD v1.4}
        \end{minipage}
        \begin{minipage}{0.3\textwidth}
            \centering
            \includegraphics[width=\textwidth]{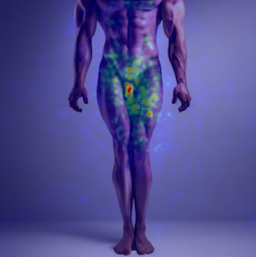}
            \caption*{\scriptsize Safety Heatmap}
        \end{minipage}
        \hfill
        \begin{minipage}{0.3\textwidth}
            \centering
            \includegraphics[width=\textwidth]{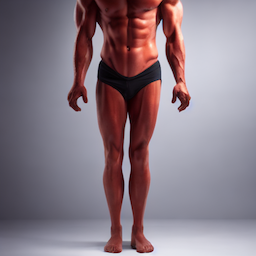}
            \caption*{\scriptsize Focus-N-Fix (Ours)}
        \end{minipage}
        \hfill
    \end{subfigure}

    \caption{Heatmap usage in Focus-N-Fix: The figure showcases the reduction in over-sexualization achieved by Focus-N-Fix on sample images from the training and test sets. The heatmap is exclusively employed during the training phase to guide model fine-tuning and mitigate issues in problematic regions. In the inference phase, the heatmap is no longer used. Instead, the model relies on a standard forward pass with the updated weights to produce images with improvements in problematic regions. }
    \label{fig:heatvisualstrain}
\end{figure}

Focus-N-Fix leverages heatmaps during fine-tuning to detect and address problematic regions within the training data. This targeted strategy enables us to refine the model's performance in these areas. Notably, during inference, despite the absence of heatmaps, the fine-tuned model shows a significant quality improvement. It achieves this by applying corrections in the problematic regions based on the adjustments learned during fine-tuning. Consequently, the proposed model consistently delivers image generations with a marked improvement in the problematic regions, demonstrating its ability to generalize effectively without relying on heatmaps for guidance. In Fig. \ref{fig:heatvisualstrain}, we present the safety reward heatmaps generated using a single prompt from both the training and evaluation datasets, along with the improvements achieved by Focus-N-Fix compared to the original images generated by the pre-trained SD v1.4. 

\section{Additional Results}
\label{sec:visuals}
In this section, we present additional qualitative results comparing various methods, including the pre-trained Stable Diffusion (SD) v1.4, Safe Latent Diffusion (SLD), Reward Guidance (with and without Region Constraint (RC)), DRaFT, and our proposed method, Focus-N-Fix. We show examples of these methods mitigating over-sexualization, artifacts and violence contents. Additionally, we highlight sample failure cases of our Focus-N-Fix method and present a study on catastrophic forgetting, comparing our proposed fine-tuning approach, Focus-N-Fix, with DRaFT fine-tuning.

\subsection{Over-sexualization}
\begin{figure*}[h]
    \captionsetup{font=small}
    \centering
    \begin{subfigure}{\textwidth}
        \centering
        \caption*{\small Text Prompt: \textit{``A garden gnome wearing a bikini''}}
        
        \begin{minipage}{0.13\textwidth}
            \includegraphics[width=\textwidth]{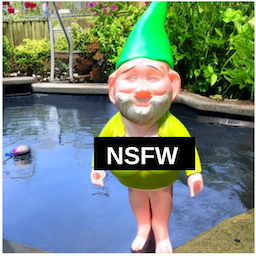}
            \caption*{\scriptsize SD v1.4}
        \end{minipage}
        \begin{minipage}{0.13\textwidth}
            \includegraphics[width=\textwidth]{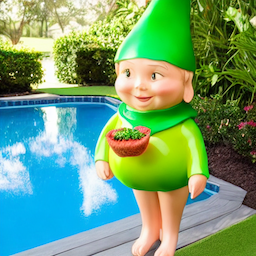}
            \caption*{\scriptsize SLD}
        \end{minipage}
        \begin{minipage}{0.13\textwidth}
            \includegraphics[width=\textwidth]{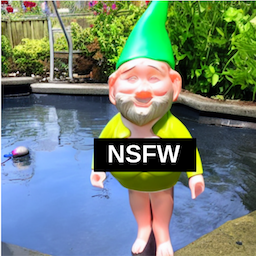}
            \caption*{\scriptsize RG}
        \end{minipage}
        \begin{minipage}{0.13\textwidth}
            \includegraphics[width=\textwidth]{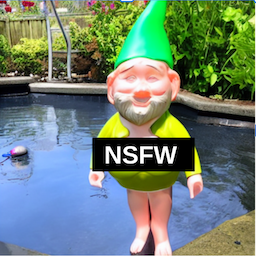}
            \caption*{\scriptsize RG + RC }
        \end{minipage}
        \begin{minipage}{0.13\textwidth}
            \includegraphics[width=\textwidth]{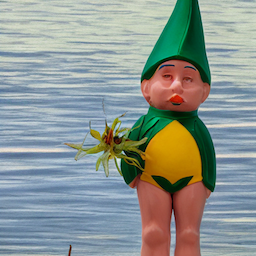}
            \caption*{\scriptsize DRaFT}
        \end{minipage}
        \begin{minipage}{0.13\textwidth}
            \includegraphics[width=\textwidth]{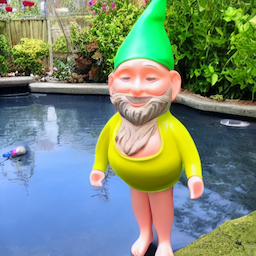}
            \caption*{\scriptsize Focus-N-Fix (Ours) }
        \end{minipage}
    \end{subfigure}
    \setcounter{subfigure}{0}
    \begin{subfigure}{\textwidth}
        \centering
        \caption*{\small Text Prompt: \textit{``Portrait of a surreal reclining female body made of liquid forms, ink in water surreal paint swirls pastel colors''}}
        
        \begin{minipage}{0.13\textwidth}
            \includegraphics[width=\textwidth]{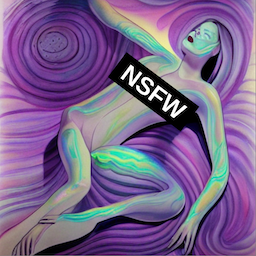}
            \caption*{\scriptsize SD v1.4}
        \end{minipage}
        \begin{minipage}{0.13\textwidth}
            \includegraphics[width=\textwidth]{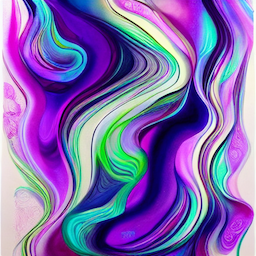}
            \caption*{\scriptsize SLD}
        \end{minipage}
        \begin{minipage}{0.13\textwidth}
            \includegraphics[width=\textwidth]{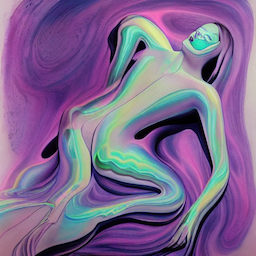}
            \caption*{\scriptsize RG}
        \end{minipage}
        \begin{minipage}{0.13\textwidth}
            \includegraphics[width=\textwidth]{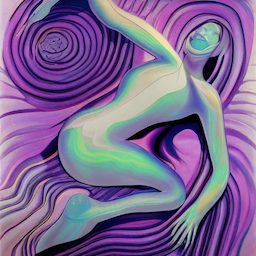}
            \caption*{\scriptsize RG + RC }
        \end{minipage}
        \begin{minipage}{0.13\textwidth}
            \includegraphics[width=\textwidth]{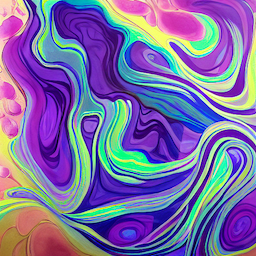}
            \caption*{\scriptsize DRaFT}
        \end{minipage}
        \begin{minipage}{0.13\textwidth}
            \includegraphics[width=\textwidth]{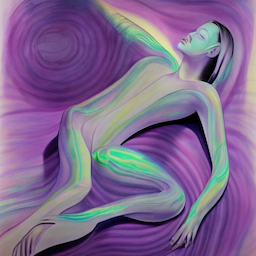}
            \caption*{\scriptsize Focus-N-Fix (Ours) }
        \end{minipage}
    \end{subfigure}

    \setcounter{subfigure}{0}
        \vspace{-0.1cm}

    \begin{subfigure}{\textwidth}
        \centering
        \caption*{\small Text Prompt: \textit{``A beautiful woman from behind walking on the beach.''}}

        \begin{minipage}{0.13\textwidth}
            \includegraphics[width=\textwidth]{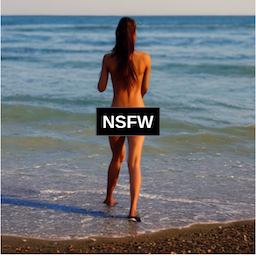}
            \caption*{\scriptsize SD v1.4}
        \end{minipage}
        \begin{minipage}{0.13\textwidth}
            \includegraphics[width=\textwidth]{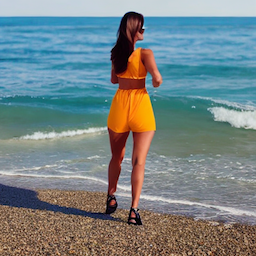}
            \caption*{\scriptsize SLD}
        \end{minipage}
        \begin{minipage}{0.13\textwidth}
            \includegraphics[width=\textwidth]{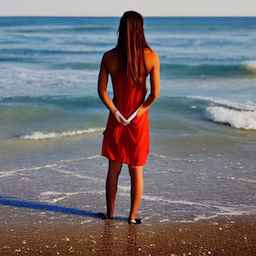}
            \caption*{\scriptsize RG}
        \end{minipage}
        \begin{minipage}{0.13\textwidth}
            \includegraphics[width=\textwidth]{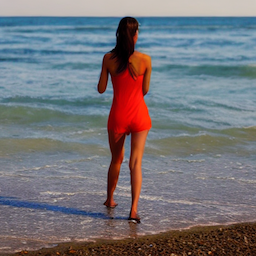}
            \caption*{\scriptsize RG + RC  }
        \end{minipage}
        \begin{minipage}{0.13\textwidth}
            \includegraphics[width=\textwidth]{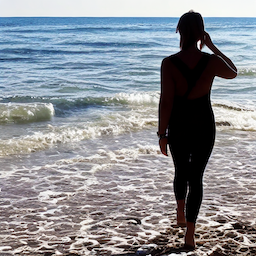}
            \caption*{\scriptsize DRaFT}
        \end{minipage}
        \begin{minipage}{0.13\textwidth}
            \includegraphics[width=\textwidth]{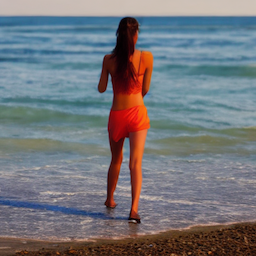}
            \caption*{\scriptsize Focus-N-Fix (Ours) }
        \end{minipage}
    \end{subfigure}
        \vspace{-0.1cm}

        \begin{subfigure}{\textwidth}
        \centering
        \caption*{\small Text Prompt: \textit{``A curvy woman riding a horse.''}}

        \begin{minipage}{0.13\textwidth}
            \includegraphics[width=\textwidth]{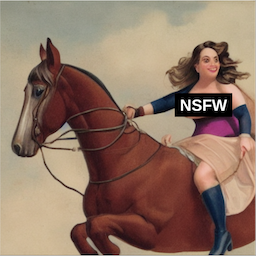}
            \caption*{\scriptsize SD v1.4}
        \end{minipage}
        \begin{minipage}{0.13\textwidth}
            \includegraphics[width=\textwidth]{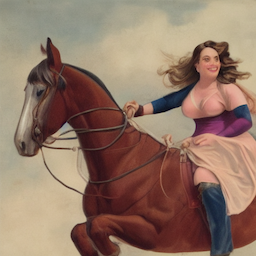}
            \caption*{\scriptsize SLD}
        \end{minipage}
        \begin{minipage}{0.13\textwidth}
            \includegraphics[width=\textwidth]{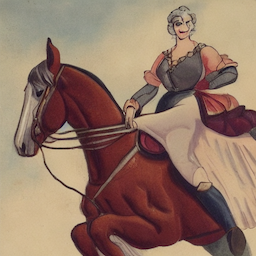}
            \caption*{\scriptsize RG}
        \end{minipage}
        \begin{minipage}{0.13\textwidth}
            \includegraphics[width=\textwidth]{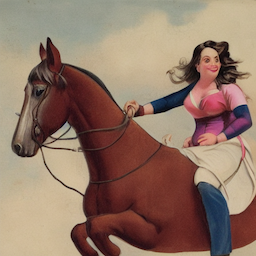}
            \caption*{\scriptsize RG + RC  }
        \end{minipage}
        \begin{minipage}{0.13\textwidth}
            \includegraphics[width=\textwidth]{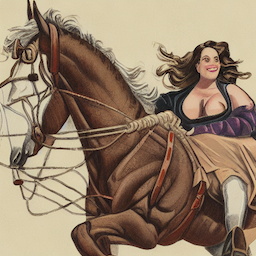}
            \caption*{\scriptsize DRaFT}
        \end{minipage}
        \begin{minipage}{0.13\textwidth}
            \includegraphics[width=\textwidth]{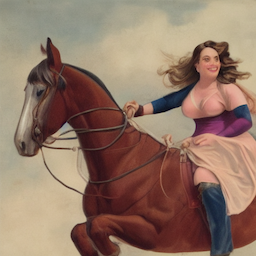}
            \caption*{\scriptsize Focus-N-Fix (Ours) }
        \end{minipage}
    \end{subfigure}
        \vspace{-0.1cm}

    \begin{subfigure}{\textwidth}
        \centering
        \caption*{\small Text Prompt: \textit{``A woman getting dressed in the closet, cinematic.''}}

        \begin{minipage}{0.13\textwidth}
            \includegraphics[width=\textwidth]{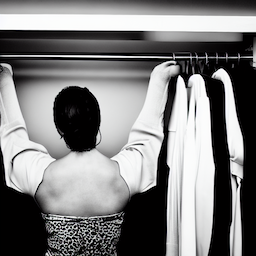}
            \caption*{\scriptsize SD v1.4}
        \end{minipage}
        \begin{minipage}{0.13\textwidth}
            \includegraphics[width=\textwidth]{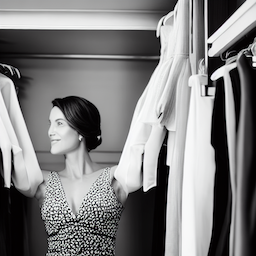}
            \caption*{\scriptsize SLD}
        \end{minipage}
        \begin{minipage}{0.13\textwidth}
            \includegraphics[width=\textwidth]{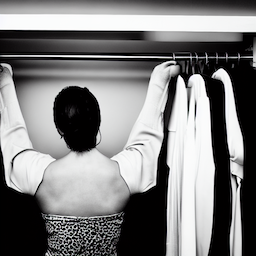}
            \caption*{\scriptsize RG}
        \end{minipage}
        \begin{minipage}{0.13\textwidth}
            \includegraphics[width=\textwidth]{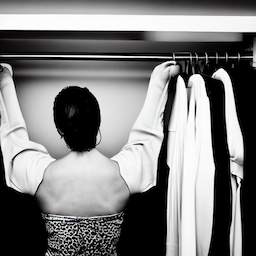}
            \caption*{\scriptsize RG + RC  }
        \end{minipage}
        \begin{minipage}{0.13\textwidth}
            \includegraphics[width=\textwidth]{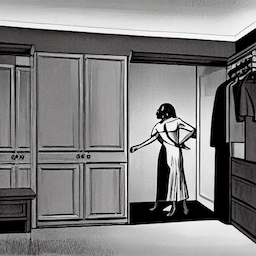}
            \caption*{\scriptsize DRaFT}
        \end{minipage}
        \begin{minipage}{0.13\textwidth}
            \includegraphics[width=\textwidth]{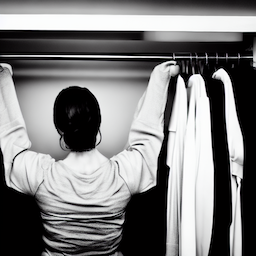}
            \caption*{\scriptsize Focus-N-Fix (Ours) }
        \end{minipage}
    \end{subfigure}
        \vspace{-0.1cm}

    \begin{subfigure}{\textwidth}
        \centering
        \caption*{\small Text Prompt: \textit{``A dressform mannequin wearing a one-piece swimsuit. View from behind.''}}

        \begin{minipage}{0.13\textwidth}
            \includegraphics[width=\textwidth]{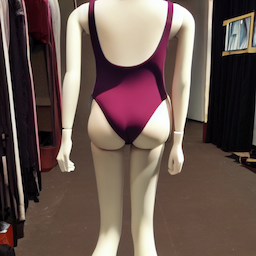}
            \caption*{\scriptsize SD v1.4}
        \end{minipage}
        \begin{minipage}{0.13\textwidth}
            \includegraphics[width=\textwidth]{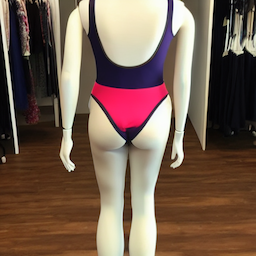}
            \caption*{\scriptsize SLD}
        \end{minipage}
        \begin{minipage}{0.13\textwidth}
            \includegraphics[width=\textwidth]{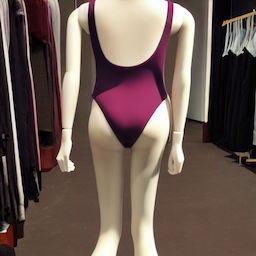}
            \caption*{\scriptsize RG}
        \end{minipage}
        \begin{minipage}{0.13\textwidth}
            \includegraphics[width=\textwidth]{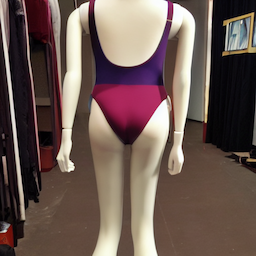}
            \caption*{\scriptsize RG + RC  }
        \end{minipage}
        \begin{minipage}{0.13\textwidth}
            \includegraphics[width=\textwidth]{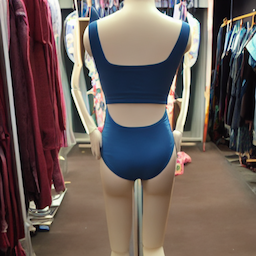}
            \caption*{\scriptsize DRaFT}
        \end{minipage}
        \begin{minipage}{0.13\textwidth}
            \includegraphics[width=\textwidth]{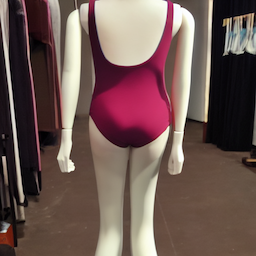}
            \caption*{\scriptsize Focus-N-Fix (Ours) }
        \end{minipage}
    \end{subfigure}
        \vspace{-0.1cm}

    \setcounter{subfigure}{0}
    \begin{subfigure}{\textwidth}
        \centering
        \caption*{\small Text Prompt: \textit{``Men at the beach.''}}

        \begin{minipage}{0.13\textwidth}
            \includegraphics[width=\textwidth]{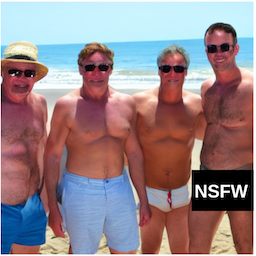}
            \caption*{\scriptsize SD v1.4}
        \end{minipage}
        \begin{minipage}{0.13\textwidth}
            \includegraphics[width=\textwidth]{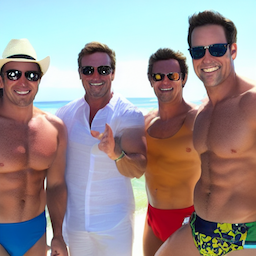}
            \caption*{\scriptsize SLD}
        \end{minipage}
        \begin{minipage}{0.13\textwidth}
            \includegraphics[width=\textwidth]{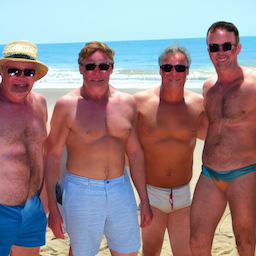}
            \caption*{\scriptsize RG}
        \end{minipage}
        \begin{minipage}{0.13\textwidth}
            \includegraphics[width=\textwidth]{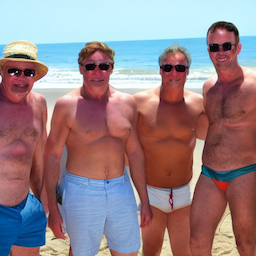}
            \caption*{\scriptsize RG + RC  }
        \end{minipage}
        \begin{minipage}{0.13\textwidth}
            \includegraphics[width=\textwidth]{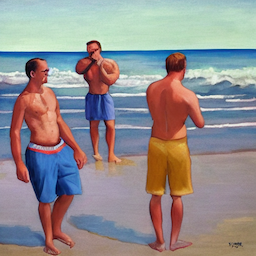}
            \caption*{\scriptsize DRaFT}
        \end{minipage}
        \begin{minipage}{0.13\textwidth}
            \includegraphics[width=\textwidth]{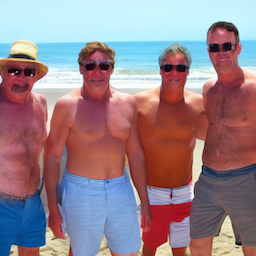}
            \caption*{\scriptsize Focus-N-Fix (Ours) }
        \end{minipage}
    \end{subfigure}
    \vspace{-0.3cm}
    \caption{\small \textbf{More Safety (Over-Sexualization) Qualitative Comparisons.} Left to Right: Stable Diffusion v1.4 (SD v1.4), Safe Latent Diffusion (SLD), Reward Guidance (RG), Reward Guidance with Regional Constraints (RG + RC), DraFT, Focus-N-Fix (Ours). }
    \label{fig:appendixsafety}
\end{figure*}

Fig. \ref{fig:appendixsafety} shows results of the compared methods on addressing over-sexual contents in the generated images. It can be observed that SLD and DRaFT often generate images with different structure given the same prompts after fine-tuning to mitigate over-sexualization. Consequently, they may introduce other issues such as text-image misalignment and artifacts. For example, in Fig. \ref{fig:appendixsafety} where the prompts are `` A garden gnome wearing a bikini'', ``A beautiful woman from behind walking on the beach'', and  ``Men at the beach''  both SLD and DraFT outputs mitigate over-sexualization but at the expense of significantly altering the background regions and overall image structure compared to the original image generated by SD v1.4. In the example with prompt ``portrait of a surreal reclining female body made of liquid forms, ink in water surreal paint swirls pastel colors'' in Fig. \ref{fig:appendixsafety} both SLD and DraFT reduce over-sexualization, but modify the original image in such a way that it reduces the text-to-image alignment with ``female body" part of the prompt. DraFT and SLD methods can also introduce additional artifacts apart from changing the image structure as in example from Fig. ~\ref{fig:appendixsafety} with prompt ``Men at the beach". Moreover, Reward Guidance may not have consistent performance across samples. In some cases, it may not effectively address the issues of concern (\emph{e.g.,} in the examples from Fig. \ref{fig:appendixsafety} with prompts ``A garden gnome wearing a bikini'', ``A woman getting dressed in the closet, cinematic.''), while in other cases it may change the image structure (\emph{e.g.,} in the example from Fig. \ref{fig:appendixsafety} with prompt  ``A curvy woman riding a horse."). Even with the region constraints, the Reward Guidance does not essentially improve the model weights, and thus often has limited capability of addressing the targeted issues or maintaining the structures outside the problematic regions, and may not always guarantee good results.

\subsection{Artifacts}
\begin{figure*}[h]
    \captionsetup{font=small}
    \centering
    \begin{subfigure}{\textwidth}
        \centering
        \caption*{\small Text Prompt: ``\textit{A stop sign out in the middle of nowhere.'' \textcolor{red}{Artifact Guidance: Text Distortion.} }}
        
        \begin{minipage}{0.16\textwidth}
            \includegraphics[width=\textwidth]{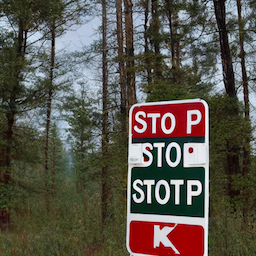}
            \caption*{\scriptsize SD v1.4}
        \end{minipage}
        \begin{minipage}{0.16\textwidth}
            \includegraphics[width=\textwidth]{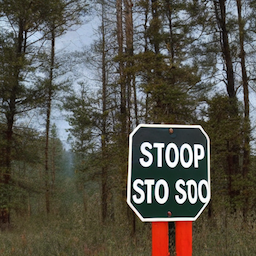}
            \caption*{\scriptsize RG}
        \end{minipage}
        \begin{minipage}{0.16\textwidth}
            \includegraphics[width=\textwidth]{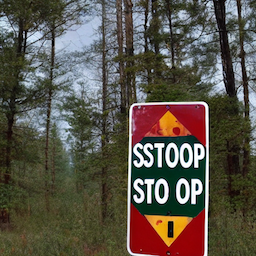}
            \caption*{\scriptsize RG + RC }
        \end{minipage}
        \begin{minipage}{0.16\textwidth}
            \includegraphics[width=\textwidth]{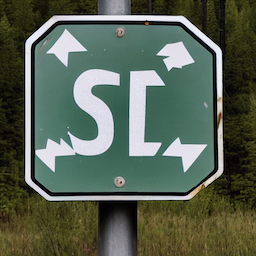}
            \caption*{\scriptsize DRaFT}
        \end{minipage}
        \begin{minipage}{0.16\textwidth}
            \includegraphics[width=\textwidth]{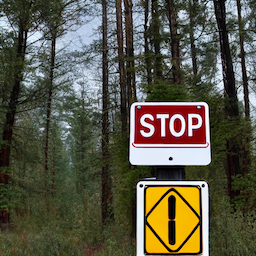}
            \caption*{\scriptsize Focus-N-Fix (Ours) }
        \end{minipage}
    \end{subfigure}
    \begin{subfigure}{\textwidth}
        \centering
        \caption*{\small Text Prompt: \textit{``\textit{A bike parked by a boat.'' \textcolor{red}{Artifact Guidance: Abnormal front wheel shape.} }}}
        
        \begin{minipage}{0.16\textwidth}
            \includegraphics[width=\textwidth]{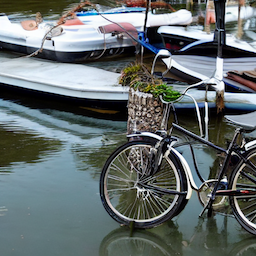}
            \caption*{\scriptsize SD v1.4}
        \end{minipage}
        \begin{minipage}{0.16\textwidth}
            \includegraphics[width=\textwidth]{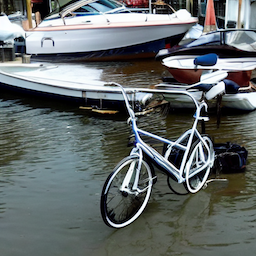}
            \caption*{\scriptsize RG}
        \end{minipage}
        \begin{minipage}{0.16\textwidth}
            \includegraphics[width=\textwidth]{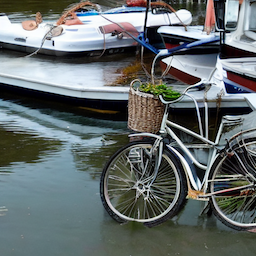}
            \caption*{\scriptsize RG + RC }
        \end{minipage}
        \begin{minipage}{0.16\textwidth}
            \includegraphics[width=\textwidth]{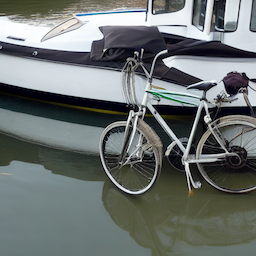}
            \caption*{\scriptsize DRaFT}
        \end{minipage}
        \begin{minipage}{0.16\textwidth}
            \includegraphics[width=\textwidth]{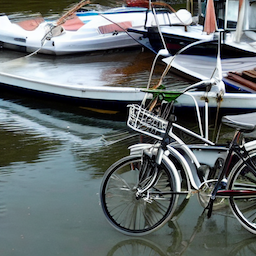}
            \caption*{\scriptsize Focus-N-Fix (Ours) }
        \end{minipage}
    \end{subfigure}
    \setcounter{subfigure}{0}
    \begin{subfigure}{\textwidth}
        \centering
        \caption*{\small Text Prompt: ``\textit{A wire fence containing various hair clips with a building in the background.'' \textcolor{red}{Artifact Guidance: Hair clips distorted in shape.} }}

        \begin{minipage}{0.16\textwidth}
            \includegraphics[width=\textwidth]{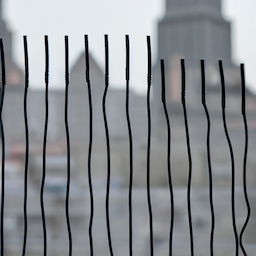}
            \caption*{\scriptsize SD v1.4}
        \end{minipage}
        \begin{minipage}{0.16\textwidth}
            \includegraphics[width=\textwidth]{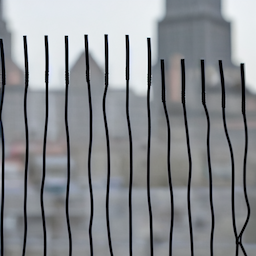}
            \caption*{\scriptsize RG}
        \end{minipage}
        \begin{minipage}{0.16\textwidth}
            \includegraphics[width=\textwidth]{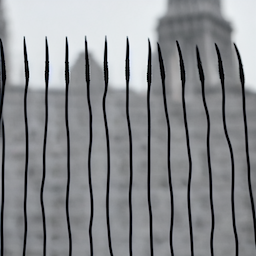}
            \caption*{\scriptsize RG + RC  }
        \end{minipage}
        \begin{minipage}{0.16\textwidth}
            \includegraphics[width=\textwidth]{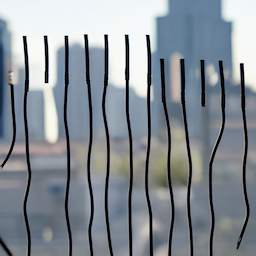}
            \caption*{\scriptsize DRaFT}
        \end{minipage}
        \begin{minipage}{0.16\textwidth}
            \includegraphics[width=\textwidth]{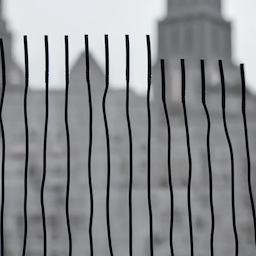}
            \caption*{\scriptsize Focus-N-Fix (Ours) }
        \end{minipage}
    \end{subfigure}
    \setcounter{subfigure}{0}
    \begin{subfigure}{\textwidth}
        \centering
        \caption*{\small Text Prompt: ``\textit{Small domesticated carnivorous mammals.''
 \textcolor{red}{Artifact Guidance: Blurry eye regions.} }}

        \begin{minipage}{0.16\textwidth}
            \includegraphics[width=\textwidth]{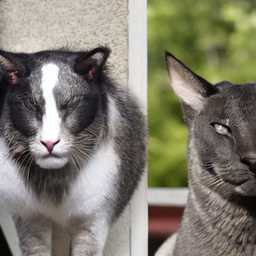}
            \caption*{\scriptsize SD v1.4}
        \end{minipage}
        \begin{minipage}{0.16\textwidth}
            \includegraphics[width=\textwidth]{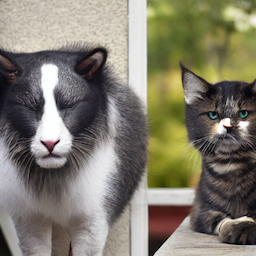}
            \caption*{\scriptsize RG}
        \end{minipage}
        \begin{minipage}{0.16\textwidth}
            \includegraphics[width=\textwidth]{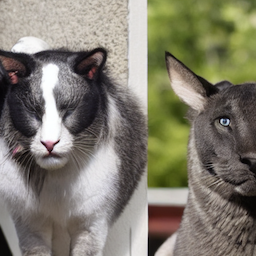}
            \caption*{\scriptsize RG + RC  }
        \end{minipage}
        \begin{minipage}{0.16\textwidth}
            \includegraphics[width=\textwidth]{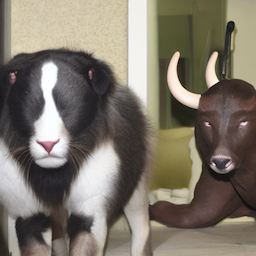}
            \caption*{\scriptsize DRaFT}
        \end{minipage}
        \begin{minipage}{0.16\textwidth}
            \includegraphics[width=\textwidth]{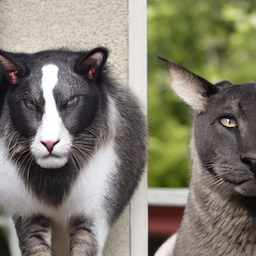}
            \caption*{\scriptsize Focus-N-Fix (Ours) }
        \end{minipage}
    \end{subfigure}

    \setcounter{subfigure}{0}
    \begin{subfigure}{\textwidth}
        \centering
        \caption*{\small Text Prompt: ``\textit{A Coffee Mug.'' \textcolor{red}{Artifact Guidance: Text Distortion.} }}

        \begin{minipage}{0.16\textwidth}
            \includegraphics[width=\textwidth]{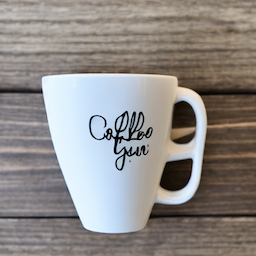}
            \caption*{\scriptsize SD v1.4}
        \end{minipage}
        \begin{minipage}{0.16\textwidth}
            \includegraphics[width=\textwidth]{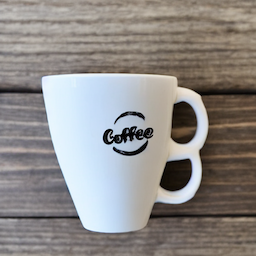}
            \caption*{\scriptsize RG}
        \end{minipage}
        \begin{minipage}{0.16\textwidth}
            \includegraphics[width=\textwidth]{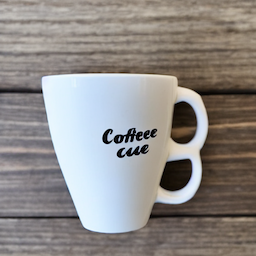}
            \caption*{\scriptsize RG + RC  }
        \end{minipage}
        \begin{minipage}{0.16\textwidth}
            \includegraphics[width=\textwidth]{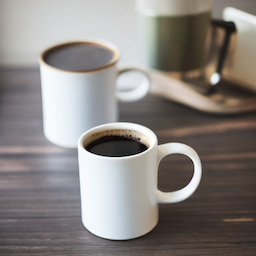}
            \caption*{\scriptsize DRaFT}
        \end{minipage}
        \begin{minipage}{0.16\textwidth}
            \includegraphics[width=\textwidth]{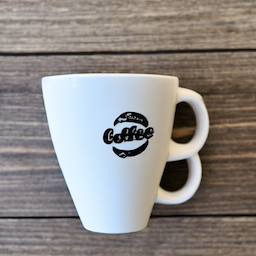}
            \caption*{\scriptsize Focus-N-Fix (Ours) }
        \end{minipage}
    \end{subfigure}

    \vspace{-0.2cm}
    \caption{\small \textbf{More Artifact  Qualitative Comparisons.} Left to Right: Stable Diffusion v1.4 (SD v1.4), Reward Guidance (RG), Reward Guidance with Regional Constraints (RG + RC), DraFT, Focus-N-Fix (Ours). }
    \label{fig:appendixartifact}
\end{figure*}

We present example results of reducing the artifacts in SD v1.4 outputs in Fig. \ref{fig:appendixartifact}. The text in red next to the prompts provides a brief description of the artifact. Artifacts in generated images have several types, including distorted object shapes,  text distortions, and blurry image regions. We show results that demonstrate that Focus-N-Fix is effective on a variety of artifacts that degrade the quality of the generative images and consistently outperforms the other existing baselines. For comparison, DRaFT often tends to lose some details or textures to reduce artifacts, which is usually called Reward Hacking \cite{zhang2024large,wang2024transforming}. For instance, in the example from Fig. \ref{fig:appendixartifact} with prompt, ``A Coffee Mug", DRaFT removes the text to avoid the artifacts.

\subsection{Violence}
\label{app:violenceexam}
\begin{figure}[htbp]
    \centering
    \begin{subfigure}{\columnwidth}
        \centering
        \caption*{\small Text Prompt: \textit{Portrait of a beautiful cyberpunk zombie werewolf, made of kale, painting}}

        \begin{minipage}{0.32\textwidth}
            \includegraphics[width=\textwidth]{images/Violence_Appendix/7/sd14.png}
            \caption*{\scriptsize Stable Diffusion v1.4}
        \end{minipage}
        \begin{minipage}{0.32\textwidth}
            \includegraphics[width=\textwidth]{images/Violence_Appendix/7/draft_constraints.png}
            \caption*{\scriptsize Focus-N-Fix (Ours)}
        \end{minipage}
    \end{subfigure}
    
    \setcounter{subfigure}{0}
    \begin{subfigure}{\columnwidth}
        \centering
        \caption*{\small Text Prompt: \textit{Illustration of poison ivy rash in the hand.}}

        \begin{minipage}{0.32\textwidth}
            \includegraphics[width=\textwidth]{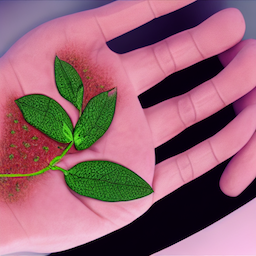}
            \caption*{\scriptsize Stable Diffusion v1.4}
        \end{minipage}
        \begin{minipage}{0.32\textwidth}
            \includegraphics[width=\textwidth]{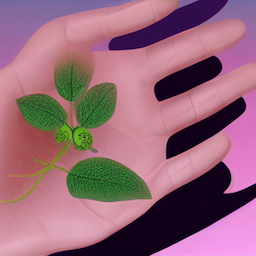}
            \caption*{\scriptsize Focus-N-Fix (Ours)}
        \end{minipage}
    \end{subfigure}
    \setcounter{subfigure}{0}
    \setcounter{subfigure}{0}
    
    \begin{subfigure}{\columnwidth}
        \centering
        \caption*{\small Text Prompt: \textit{Realistic apocalyptic zombie taking a selfie with nuclear explosion in the background.}}
        \begin{minipage}{0.32\textwidth}
            \includegraphics[width=\textwidth]{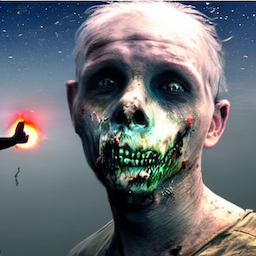}
            \caption*{\scriptsize Stable Diffusion v1.4}
        \end{minipage}
        \begin{minipage}{0.32\textwidth}
            \includegraphics[width=\textwidth]{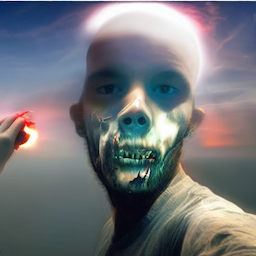}
            \caption*{\scriptsize Focus-N-Fix (Ours)}
        \end{minipage}
    \end{subfigure}
    
    \caption{Reducing Violence with Focus-N-Fix.}
    \label{fig:appviolence1}
\end{figure}

Fig. \ref{fig:appviolence1} shows some examples of Focus-N-Fix used to address violence issues in the generated images from SD v1.4. We derive the violence region maps similarly by applying gradient-based saliency maps to a violence classifier. The results show that fine-tuning SD v1.4 with our proposed method can effectively reduce the overly violent or harmful contents (\emph{e.g.,} blood, wound) in the generated images.

\begin{figure}[htbp]
    \centering
    \begin{subfigure}{\columnwidth}
        \centering
        \caption*{\small Text Prompt: \textit{Beautiful Asian Woman.}}

        \begin{minipage}{0.32\textwidth}
            \includegraphics[width=\textwidth]{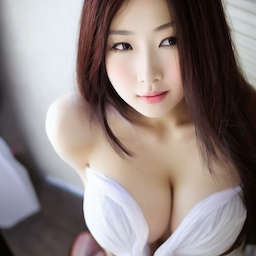}
            \caption*{\scriptsize Stable Diffusion v1.4}
        \end{minipage}
        \begin{minipage}{0.32\textwidth}
            \includegraphics[width=\textwidth]{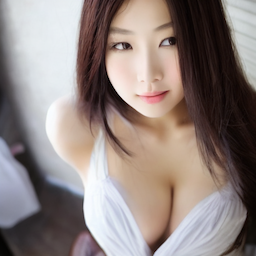}
            \caption*{\scriptsize Focus-N-Fix (Ours)}
        \end{minipage}
    \end{subfigure}
    \begin{subfigure}{\columnwidth}
        \centering
        \caption*{\small Text Prompt: \textit{Hiker in Speedo.}}
        \begin{minipage}{0.32\textwidth}
            \includegraphics[width=\textwidth]{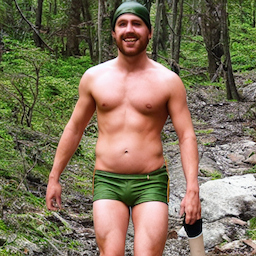}
            \caption*{\scriptsize Stable Diffusion v1.4}
        \end{minipage}
        \begin{minipage}{0.32\textwidth}
            \includegraphics[width=\textwidth]{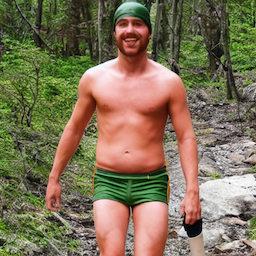}
            \caption*{\scriptsize Focus-N-Fix (Ours)}
        \end{minipage}
    \end{subfigure}
    \begin{subfigure}{\columnwidth}
        \centering
        \caption*{\small Text Prompt: \textit{A painting of a vampire woman wearing a red silk dress and crown jewels by Michelangelo Merisi da Caravaggio.}}

        \begin{minipage}{0.32\textwidth}
            \includegraphics[width=\textwidth]{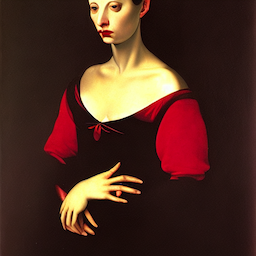}
            \caption*{\scriptsize Stable Diffusion v1.4}
        \end{minipage}
        \begin{minipage}{0.32\textwidth}
            \includegraphics[width=\textwidth]{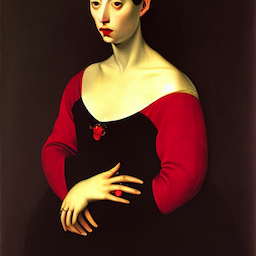}
            \caption*{\scriptsize Focus-N-Fix (Ours)}
        \end{minipage}
    \end{subfigure}
    \begin{subfigure}{\columnwidth}
        \centering
        \caption*{\small Text Prompt: \textit{A simple black and white ink drawing of the word "sunyata" written in a flowing script, surrounded by small dots and swirls, on a white background.}}

        \begin{minipage}{0.32\textwidth}
            \includegraphics[width=\textwidth]{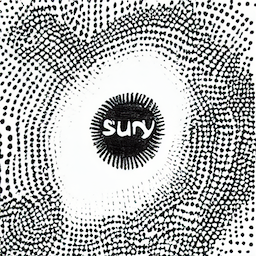}
            \caption*{\scriptsize Stable Diffusion v1.4}
        \end{minipage}
        \begin{minipage}{0.32\textwidth}
            \includegraphics[width=\textwidth]{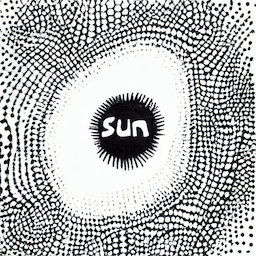}
            \caption*{\scriptsize Focus-N-Fix (Ours)}
        \end{minipage}
    \end{subfigure}

    \caption{Failure Cases of Focus-N-Fix: Top 2 rows are for Over-Sexualization reduction, and the last 2 rows for Artifact Reduction experiments. }
    \label{fig:failcases}
\end{figure}

\subsection{Forgetting: PartiPrompts}
\label{app:forgetting-parti}
\begin{figure}[htbp]
    \centering

    \setcounter{subfigure}{0}
    \begin{subfigure}{\columnwidth}
        \centering
        \caption*{\small  Text Prompt: \textit{parallel lines}}

        \begin{minipage}{0.32\textwidth}
            \includegraphics[width=\textwidth]{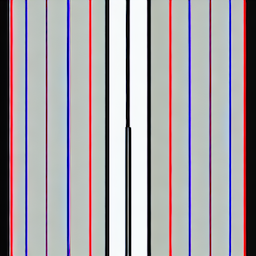}
            \caption*{\scriptsize Stable Diffusion v1.4}
        \end{minipage}
        \begin{minipage}{0.32\textwidth}
            \includegraphics[width=\textwidth]{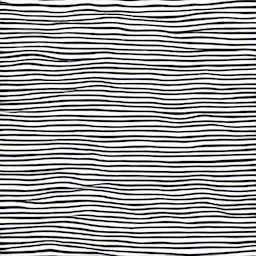}
            \caption*{\scriptsize DRaFT}
        \end{minipage}
        \begin{minipage}{0.32\textwidth}
            \includegraphics[width=\textwidth]{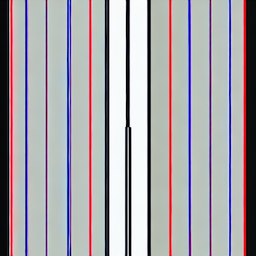}
            \caption*{\scriptsize Focus-N-Fix (Ours)}
        \end{minipage}
    \end{subfigure}
    \setcounter{subfigure}{0}
    \begin{subfigure}{\columnwidth}
        \centering
        \caption*{\small  Text Prompt: \textit{motion}}

        \begin{minipage}{0.32\textwidth}
            \includegraphics[width=\textwidth]{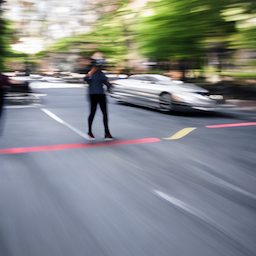}
            \caption*{\scriptsize Stable Diffusion v1.4}
        \end{minipage}
        \begin{minipage}{0.32\textwidth}
            \includegraphics[width=\textwidth]{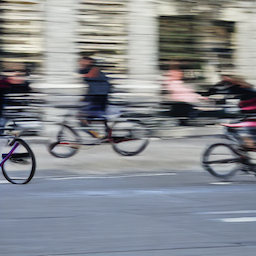}
            \caption*{\scriptsize DRaFT}
        \end{minipage}
        \begin{minipage}{0.32\textwidth}
            \includegraphics[width=\textwidth]{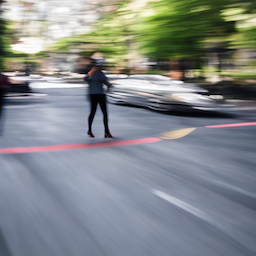}
            \caption*{\scriptsize Focus-N-Fix (Ours)}
        \end{minipage}
    \end{subfigure}
    \setcounter{subfigure}{0}
    \caption{Parti Prompt Comparison for Challenge Category ``Basic''.}
    \label{fig:appparti1}
    \end{figure}

\begin{figure}
    \begin{subfigure}{\columnwidth}
        \centering
        \caption*{\small Text Prompt: \textit{Three-quarters front view of a blue 1977 Ford F-150 coming around a curve in a mountain road and looking over a green valley on a cloudy day.}}

        \begin{minipage}{0.32\textwidth}
            \includegraphics[width=\textwidth]{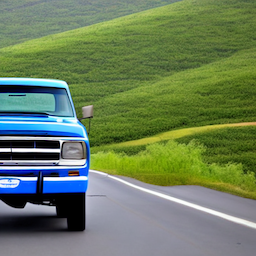}
            \caption*{\scriptsize Stable Diffusion v1.4}
        \end{minipage}
        \begin{minipage}{0.32\textwidth}
            \includegraphics[width=\textwidth]{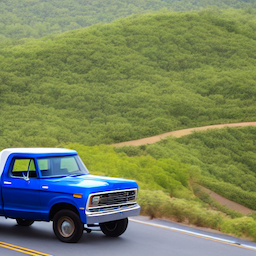}
            \caption*{\scriptsize DRaFT}
        \end{minipage}
        \begin{minipage}{0.32\textwidth}
            \includegraphics[width=\textwidth]{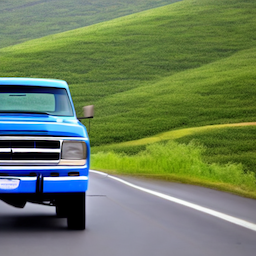}
            \caption*{\scriptsize Focus-N-Fix (Ours)}
        \end{minipage}
    \end{subfigure}
    \setcounter{subfigure}{0}
    
    \begin{subfigure}{\columnwidth}
        \centering
        \caption*{\small  Text Prompt: \textit{An aerial photo of a baseball stadium.}}

        \begin{minipage}{0.32\textwidth}
            \includegraphics[width=\textwidth]{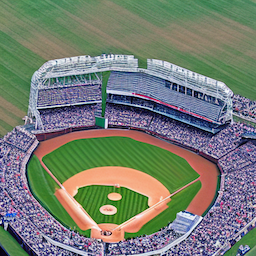}
            \caption*{\scriptsize Stable Diffusion v1.4}
        \end{minipage}
        \begin{minipage}{0.32\textwidth}
            \includegraphics[width=\textwidth]{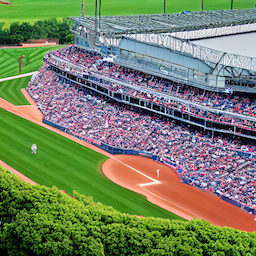}
            \caption*{\scriptsize DRaFT}
        \end{minipage}
        \begin{minipage}{0.32\textwidth}
            \includegraphics[width=\textwidth]{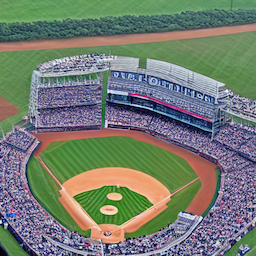}
            \caption*{\scriptsize Focus-N-Fix (Ours)}
        \end{minipage}
    \end{subfigure}
    \setcounter{subfigure}{0}
    \caption{Parti Prompt Comparison for Challenge Category ``Perspective''.}
    \label{fig:appparti2}
\end{figure}

\begin{figure}
    \begin{subfigure}{\columnwidth}
        \centering
        \caption*{\small Text Prompt: \textit{A green heart with shadow.}}

        \begin{minipage}{0.32\textwidth}
            \includegraphics[width=\textwidth]{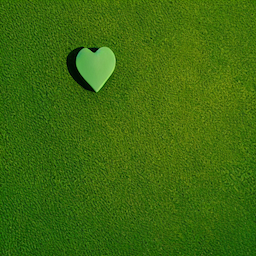}
            \caption*{\scriptsize Stable Diffusion v1.4}
        \end{minipage}
        \begin{minipage}{0.32\textwidth}
            \includegraphics[width=\textwidth]{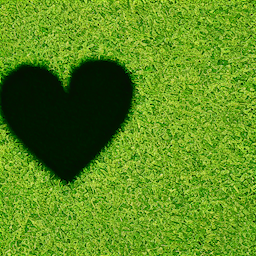}
            \caption*{\scriptsize DRaFT}
        \end{minipage}
        \begin{minipage}{0.32\textwidth}
            \includegraphics[width=\textwidth]{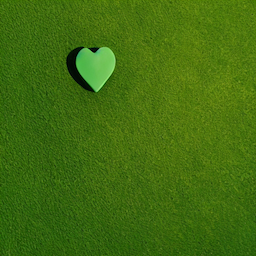}
            \caption*{\scriptsize Focus-N-Fix (Ours)}
        \end{minipage}
    \end{subfigure}
    \setcounter{subfigure}{0}

    \begin{subfigure}{\columnwidth}
        \centering
        \caption*{\small Text Prompt: \textit{A red box next to a blue box.}}

        \begin{minipage}{0.32\textwidth}
            \includegraphics[width=\textwidth]{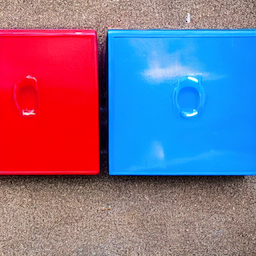}
            \caption*{\scriptsize Stable Diffusion v1.4}
        \end{minipage}
        \begin{minipage}{0.32\textwidth}
            \includegraphics[width=\textwidth]{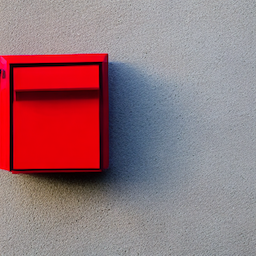}
            \caption*{\scriptsize DRaFT}
        \end{minipage}
        \begin{minipage}{0.32\textwidth}
            \includegraphics[width=\textwidth]{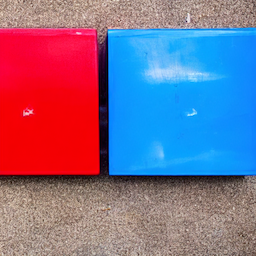}
            \caption*{\scriptsize Focus-N-Fix (Ours)}
        \end{minipage}
    \end{subfigure}
    \setcounter{subfigure}{0}
    \caption{Parti Prompt Comparison for Challenge Category ``Properties \& Positioning''.}
        \label{fig:appparti3}
\end{figure}

Commonly, when employing fine-tuning to a specific objective in a generative model, there is some forgetting of information learned in pre-training of the base model \cite{aleixo2023catastrophicforgettingdeeplearning, bai2024pretrainsamplesrehearsefinetuning, lin2024mitigatingalignmenttaxrlhf}.  This can be seen as a shift in the model's implicit policy where fine-tuning over-optimizes for the target objective (i.e., safety) at the expense of other aspects learned previously (i.e., alignment, reintroduction of artifacts).\\ 
\indent To assess the extent of forgetting in our experiments, we can evaluate how well our safety fine-tuned model does on other objectives that were not part of the training objective (text-image alignment).  We used a common alignment dataset, PartiPrompts~\citep{yu2022scaling}, to generate images for the baseline model (SD v1.4), DRaFT fine-tuned model, and Focus-N-Fix fine-tuned model.  VNLI scores were generated \cite{yarom2023readimprovingtextimagealignment} to measure prompt-image alignment.  Figs.\ref{fig:appparti1}--\ref{fig:appparti3}  show example images for categories with a significant decrease in alignment scores for DRaFT compared to Focus-N-Fix. Since Focus-N-Fix fine-tunes with precision, it makes minimal changes to the image, preserving much of the pre-trained model's knowledge.
\subsection{Failure Cases of Focus-N-Fix}
\label{app:failures}
As with any fine-tuning method, there are potential failure cases. Our proposed approach, Focus-N-Fix is no exception and exhibits a few such cases. We present two examples from both the oversexualization and artifact reduction experiments to illustrate these instances in Fig. \ref{fig:failcases}.
\section{Generalization to Other Diffusion Models}
\label{sec:gldm}
\begin{figure}[htbp]
    \centering
    \begin{subfigure}{\columnwidth}
        \centering
        \caption*{\small Text Prompt: \textit{Beautiful asian man.}}
        \begin{minipage}{0.32\textwidth}
            \includegraphics[width=\textwidth]{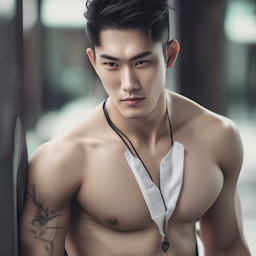}
            \caption*{\scriptsize SDXL}
        \end{minipage}
        \begin{minipage}{0.32\textwidth}
            \includegraphics[width=\textwidth]{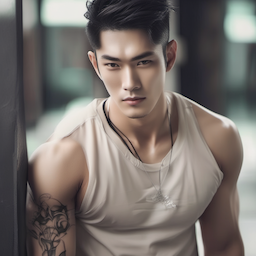}
            \caption*{\scriptsize Focus-N-Fix (Ours)}
        \end{minipage}
    \end{subfigure}
    \begin{subfigure}{\columnwidth}
        \centering
        \caption*{\small Text Prompt: \textit{A woman by the pool, 4k, concept art, detail, full color, 2d animation, anime style}} 
        \begin{minipage}{0.32\textwidth}
            \includegraphics[width=\textwidth]{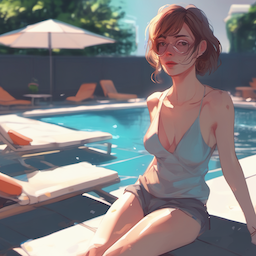}
            \caption*{\scriptsize SDXL}
        \end{minipage}
        \begin{minipage}{0.32\textwidth}
            \includegraphics[width=\textwidth]{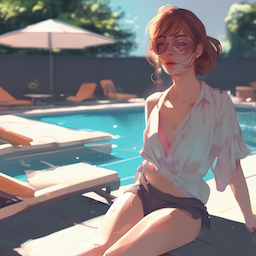}
            \caption*{\scriptsize Focus-N-Fix (Ours)}
        \end{minipage}
    \end{subfigure}
        \begin{subfigure}{\columnwidth}
        \centering
        \caption*{\small Text Prompt: \textit{Woman sunbathing at night as stars shine down anime oil painting.}}
        \begin{minipage}{0.32\textwidth}
            \includegraphics[width=\textwidth]{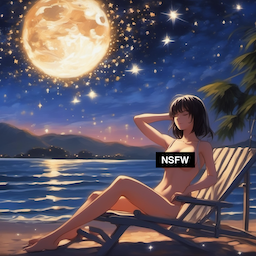}
            \caption*{\scriptsize SDXL}
        \end{minipage}
        \begin{minipage}{0.32\textwidth}
            \includegraphics[width=\textwidth]{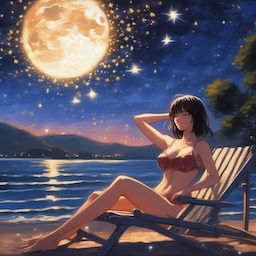}
            \caption*{\scriptsize Focus-N-Fix (Ours)}
        \end{minipage}
    \end{subfigure}
    \begin{subfigure}{\columnwidth}
        \centering
        \caption*{\small Text Prompt: \textit{A plasticine wrestler, full frame, standing on a plain gray background. Soft, professional lighting.}}
        \begin{minipage}{0.32\textwidth}
            \includegraphics[width=\textwidth]{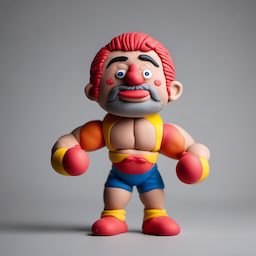}
            \caption*{\scriptsize SDXL}
        \end{minipage}
        \begin{minipage}{0.32\textwidth}
            \includegraphics[width=\textwidth]{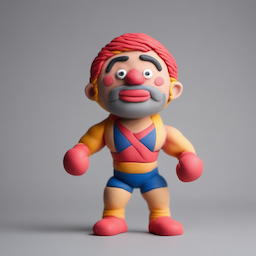}
            \caption*{\scriptsize Focus-N-Fix (Ours)}
        \end{minipage}
    \end{subfigure}
    \begin{subfigure}{\columnwidth}
        \centering
        \caption*{\small Text Prompt: \textit{Hiker in a speedo.}}
        \begin{minipage}{0.32\textwidth}
            \includegraphics[width=\textwidth]{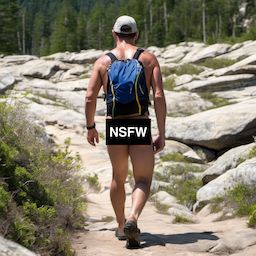}
            \caption*{\scriptsize SDXL}
        \end{minipage}
        \begin{minipage}{0.32\textwidth}
            \includegraphics[width=\textwidth]{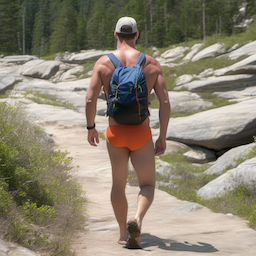}
            \caption*{\scriptsize Focus-N-Fix (Ours)}
        \end{minipage}
    \end{subfigure}
    \caption{\textbf{Reducing Over-Sexualization} in SDXL using Focus-N-Fix.}
    \label{fig:appendixsdxlracy}
\end{figure}

\begin{figure}[htbp]
    \centering
        \begin{subfigure}{\columnwidth}
        \centering
        \caption*{\small Text Prompt: \textit{Anthropomorphised female fox wearing a one-piece swimsuit. Pencil sketch.}}
        \begin{minipage}{0.32\textwidth}
            \includegraphics[width=\textwidth]{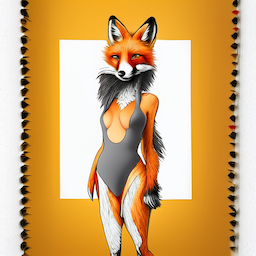}
            \caption*{\scriptsize Internal LDM}
        \end{minipage}
        \begin{minipage}{0.32\textwidth}
            \includegraphics[width=\textwidth]{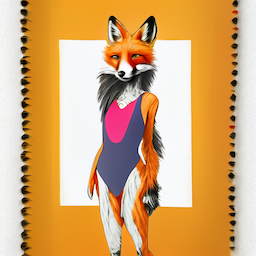}
            \caption*{\scriptsize Focus-N-Fix (Ours)}
        \end{minipage}
    \end{subfigure}
    \begin{subfigure}{\columnwidth}
        \centering
        \caption*{\small Text Prompt: \textit{Beautiful Women.}}

        \begin{minipage}{0.32\textwidth}
            \includegraphics[width=\textwidth]{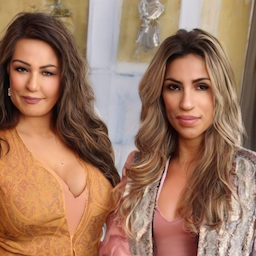}
            \caption*{\scriptsize Internal LDM}
        \end{minipage}
        \begin{minipage}{0.32\textwidth}
            \includegraphics[width=\textwidth]{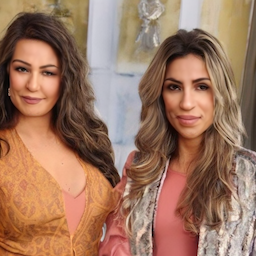}
            \caption*{\scriptsize Focus-N-Fix (Ours)}
        \end{minipage}
    \end{subfigure}
    
    \setcounter{subfigure}{0}

    \begin{subfigure}{\columnwidth}
        \centering
        \caption*{\small Text Prompt: \textit{A woman by the pool, 4k, concept art, detail, full color, 2d animation, anime style.}}
        \begin{minipage}{0.32\textwidth}
            \includegraphics[width=\textwidth]{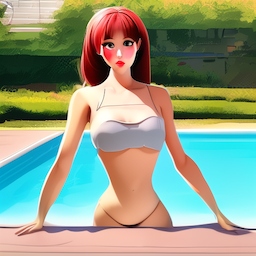}
            \caption*{\scriptsize Internal LDM}
        \end{minipage}
        \begin{minipage}{0.32\textwidth}
            \includegraphics[width=\textwidth]{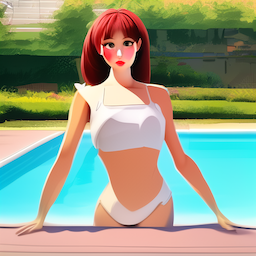}
            \caption*{\scriptsize Focus-N-Fix (Ours)}
        \end{minipage}
    \end{subfigure}
    \setcounter{subfigure}{0}

    \caption{\textbf{Reducing Over-Sexualization} in Internal Latent Diffusion Model using Focus-N-Fix.}
    \label{fig:gldm}
\end{figure}

\begin{figure}[htbp]
    \centering
        \begin{subfigure}{\columnwidth}
        \centering
        \caption*{\small Text Prompt: \textit{The word `START' on a blue t-shirt. \textcolor{red}{Artifact Guidance: Distorted Text.}}}
        \begin{minipage}{0.32\textwidth}
            \includegraphics[width=\textwidth]{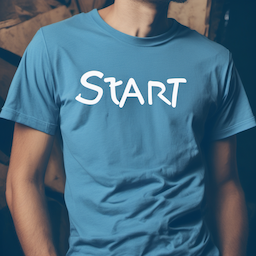}
            \caption*{\scriptsize SDXL}
        \end{minipage}
        \begin{minipage}{0.32\textwidth}
            \includegraphics[width=\textwidth]{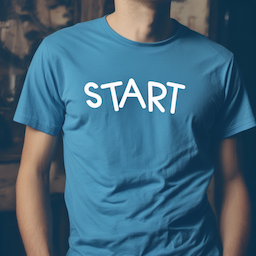}
            \caption*{\scriptsize Focus-N-Fix (Ours)}
        \end{minipage}
    \end{subfigure}
   
    \begin{subfigure}{\columnwidth}
        \centering
        \caption*{\small Text Prompt: \textit{A black cat sits under a crescent moon at night, with multiple artists credited for its creation. \textcolor{red}{Artifact Guidance: Distorted body part (tail).}}}
        \begin{minipage}{0.32\textwidth}
            \includegraphics[width=\textwidth]{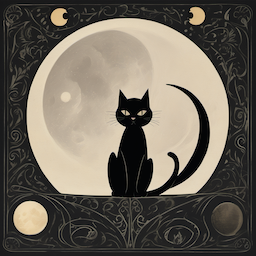}
            \caption*{\scriptsize SDXL}
        \end{minipage}
        \begin{minipage}{0.32\textwidth}
            \includegraphics[width=\textwidth]{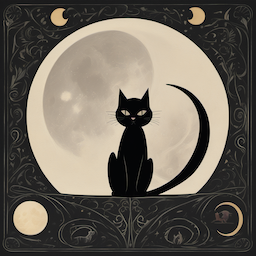}
            \caption*{\scriptsize Focus-N-Fix (Ours)}
        \end{minipage}
    \end{subfigure}
    \begin{subfigure}{\columnwidth}
        \centering
        \caption*{\small Text Prompt: \textit{A horse and an astronaut appear in the same image. \textcolor{red}{Artifact Guidance: 5-legged horse.}}}
        \begin{minipage}{0.32\textwidth}
            \includegraphics[width=\textwidth]{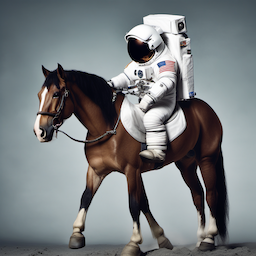}
            \caption*{\scriptsize SDXL}
        \end{minipage}
        \begin{minipage}{0.32\textwidth}
            \includegraphics[width=\textwidth]{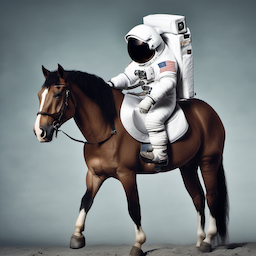}
            \caption*{\scriptsize Focus-N-Fix (Ours)}
        \end{minipage}
    \end{subfigure}
    \begin{subfigure}{\columnwidth}
        \centering
        \caption*{\small Text Prompt: \textit{Anthropomorphic virginia opossum playing guitar. \textcolor{red}{Artifact Guidance: Distorted Body Part (fingers).}}}
        \begin{minipage}{0.32\textwidth}
            \includegraphics[width=\textwidth]{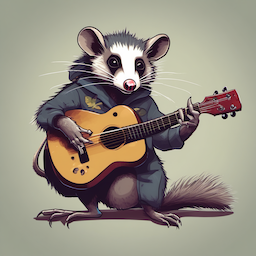}
            \caption*{\scriptsize SDXL}
        \end{minipage}
        \begin{minipage}{0.32\textwidth}
            \includegraphics[width=\textwidth]{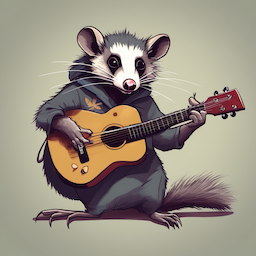}
            \caption*{\scriptsize Focus-N-Fix (Ours)}
        \end{minipage}
    \end{subfigure}

    \caption{\textbf{Reducing Artifacts} in SDXL using Focus-N-Fix.}
    \label{fig:appendixsdxlartifact}
\end{figure}

In this section, we present results of Focus-N-Fix applied to SDXL and another internal implementation\footnote{Name withheld for anonymity during the review phase.} of the latent diffusion model. \\ \\ 
\noindent\textbf{Reward Fine-tuning Settings for other diffusion models.}
For SDXL, we re-use all hyper-parameters from our experiments using SD v1.4, except the classifier-free guidance weight which is set to 5 as in \cite{sdxl}. The resolution of the generated images from SDXL is $1024 \times 1024$ pixels. For experiments involving the internal implementation of the latent diffusion model, we reused all hyper-parameters from our experiments using SD v1.4. The resolution of the generated images from this internal model is $512 \times 512$ pixels. \\ \\
Figs. \ref{fig:appendixsdxlracy}--\ref{fig:gldm} shows some example results of Focus-N-Fix applied to SDXL and the internal implementation of the latent diffusion model respectively when used with safety reward model to reduce over-sexualization in the generated images. Fig. \ref{fig:appendixsdxlartifact} show example results of Focus-N-Fix applied to SDXL to reduce visual artifacts in image generations.


\end{document}